\documentclass[11pt]{article}
\usepackage[final]{acl}
\usepackage{amsthm}
\usepackage{graphicx}
\usepackage{tabularx}
\usepackage{makecell}
\usepackage{array}
\usepackage{cuted}        
\usepackage[most]{tcolorbox}
\usepackage{amsthm}
\newtheorem{proposition}{Proposition}

\usepackage{times}
\usepackage{latexsym}
\usepackage[T1]{fontenc}
\usepackage[utf8]{inputenc}
\usepackage{microtype}
\usepackage{inconsolata}
\usepackage{graphicx}
\usepackage{subcaption}
\usepackage{booktabs}
\usepackage{amsmath,amssymb}
\usepackage{array}
\usepackage{multirow}
\usepackage{xcolor}
\usepackage{placeins}

\newcommand{\actdiff}{Activation Diffing}
\newcommand{\ourmethod}{ActDiff}
\newcommand{\domainprior}{persistent domain prior}
\newcommand{\vlm}{vision-language model}

\title{The Illusion of Expertise: Narrow Finetuning Induces Persistent Domain Priors in Vision-Language Models}

\author{
  \textbf{Qinglin Zeng\textsuperscript{1}},
  \textbf{Jusheng Zhang\textsuperscript{1}},
  \textbf{Jiangqi Huang\textsuperscript{2}},
  \textbf{Ningyuan Liu\textsuperscript{1}},
  \textbf{Luozhang Xu\textsuperscript{1}}, \\
  \textbf{Wenhao Wang\textsuperscript{3,*}},
  \textbf{Keze Wang\textsuperscript{1,*}} \\
  \textsuperscript{1}School of Computer Science and Engineering, Sun Yat-sen University \\
  \textsuperscript{2}School of Artificial Intelligence, Sun Yat-sen University
  \quad
  \textsuperscript{3}Vast Intelligence Lab \\
  \texttt{wangwenhao@vastilab.com},
  \texttt{kezewang@gmail.com}
}

\begin{document}
\maketitle

\begingroup
\renewcommand{\thefootnote}{\fnsymbol{footnote}}
\footnotetext[1]{Corresponding authors. Code is available at \url{https://github.com/zqinglin/actdiff-vlm}.}
\endgroup

\begin{abstract}
Narrow-domain finetuning makes vision-language models (VLMs) sound like domain experts while quietly making them less attentive to the image. We attribute this gap to a \emph{\domainprior{}}: an input-agnostic shift in hidden state, installed by narrow finetuning, that biases the model toward domain-style outputs regardless of visual evidence. To test the claim we introduce \ourmethod{}, a white-box diagnostic that isolates the finetuning trace by differencing base and finetuned hidden states under progressively degraded images. The trace is compact and persists under blank and Gaussian-noise inputs; the finetuned model is correspondingly \emph{less}, not more, disrupted when the image is destroyed (gap $0.323\pm0.036$ nats on \textsc{Biomed}, generic eval). Subtracting the trace at inference recovers $54\%$ of the MMStar accuracy collapse on \textsc{Biomed}, with a peak-and-collapse dose response that random and wrong-domain controls do not produce. A single first-moment regulariser added at training time shrinks the trace by $20\text{--}64\times$ across three AdaptLLM domains and improves generic NLL. The prior thus has a clean diagnostic, a causal handle at inference, and a one-line mitigation at training.

\end{abstract}

\section{Introduction}

Vision-language models are routinely adapted to specialised settings---biomedicine, document understanding, remote sensing---through narrow domain finetuning. The narrow task scores improve, but the adapted model becomes less reliable on inputs outside the finetuning distribution: it produces fluent domain prose that is only loosely anchored to the actual image. Existing accounts attribute this drift to weight-level mechanisms---erased pretrained knowledge, representation collapse, modality bias in the language backbone---and accordingly treat mitigation as a data-mixing or distillation problem.

We argue that an important part of the failure is not in the weights but in the residual stream. Narrow finetuning, in our reading, installs an \emph{\domainprior{}}: an input-agnostic shift in hidden state that the unembedding reads out as a domain-style default, regardless of what the image carries. Under this view the finetuned model behaves like a confident \emph{blind expert}---fluent in the target domain, but selectively less sensitive to visual evidence than the base model it started from. We call this the \emph{illusion of expertise}.

We turn the claim into a falsifiable diagnostic. \ourmethod{} (Section~\ref{sec:method}) differences base and finetuned hidden states under progressively degraded image inputs---shuffled, text-only, blank, and pure Gaussian noise. If the finetuning trace reflected genuine visual processing it should weaken as visual content is removed; instead, it is compact, late-layer, early-position, and just as strong on a blank image as on the original (Section~\ref{subsec:persistent-domain-prior}). On a corruption-sensitivity probe the finetuned model is correspondingly \emph{less} disturbed by destroyed images than the base---a $0.323 \pm 0.036$ nat gap on \textsc{Biomed} under generic evaluation (Section~\ref{subsec:illusion-of-expertise}). Subtracting the extracted direction at inference, without retraining, recovers $54\%$ of the \textsc{Biomed} MMStar collapse with a peak-and-collapse dose response that random-direction and wrong-domain controls do not produce (Section~\ref{subsec:causal-restoration}); the same intervention works on a second VLM family that shows the same behavioural signature, and fails on a third that does not. Finally, the prior is preventable at training: a single first-moment regulariser shrinks the diagnostic trace by $20\text{--}64\times$ across three AdaptLLM domains and improves generic NLL relative to data-mixing baselines, while a conditional KL alternative provably cannot (Section~\ref{subsec:formation-mitigation}).

The contribution is a complete loop---diagnostic, causal test, formation account, and a one-line mitigation---around the same input-agnostic mechanism, complementing rather than replacing the weight-level picture of finetuning degradation.

\section{Related Work}

\paragraph{Finetuning degradation and shortcut learning in VLMs.}
Adapting pretrained VLMs to specialized domains trades plasticity against stability~\citep{goodfellow2013,jung2023}. Narrow supervised finetuning is known to induce representation drift and to erode zero-shot out-of-distribution behavior~\citep{imanov2026,kachaev2025}, an effect tied to \emph{shortcut learning} and modality bias, where the language backbone overrides visual percepts in favor of strong linguistic priors~\citep{chi2025,nooralahzadeh2026,zheng2025,lin2024,deng2025}. Most prior accounts localize this failure to erased weights or late-stage arbitration inside the LLM. We instead characterize it as an input-agnostic activation-space prior introduced by narrow finetuning, and test its causal role directly.

\paragraph{Mechanistic interpretability and activation steering.}
Representation-engineering work treats high-level behaviors as linear directions in activation space~\citep{zou2023,tian2025}, and identifies multimodal bottlenecks such as delayed visual-entity formation that leave models vulnerable to text-prior hijacking~\citep{venhoff2025}. Extracting and injecting activation vectors has proven effective for hallucination mitigation and training-free control~\citep{yin2026,haon2025,turner2023}. We adopt this lens but use it diagnostically and causally: we isolate a finetuning-induced direction and ablate it at inference to measure its effect on standard grounding benchmarks.

\paragraph{Model diffing and activation traces.}
Isolating finetuning-induced shifts has moved from weight-space arithmetic such as task vectors~\citep{gargiulo2025,gonzalez2025,ilharco2023} to activation-space diffing~\citep{xu2025}. Closest to us is the Activation Difference Lens~\citep{minder2025}, which compares base and finetuned hidden states---building on activation inspection and intervention tools~\citep{meng2022,ghandeharioun2024}---to recover finetuning-data structure in text-only LLMs. We adapt this paradigm to VLMs as \ourmethod{}: by probing with degraded and pure Gaussian-noise images we isolate the input-agnostic modality bias and connect it to measurable grounding collapse~\citep{leng2024,wang2025}.

\section{Method}
\label{sec:method}

We define the finetuning-induced trace, motivate why it concentrates into an input-agnostic component, and specify the diagnostic and the inference-time intervention used in the experiments. Derivations are in Appendix~\ref{app:proofs}.

\subsection{Setup and Notation}
Let $f_{\text{base}}$ be a pretrained \vlm{} (vision encoder, projector, LLM backbone) and $f_{\text{ft}}$ its narrowly-finetuned counterpart on a specialised domain $\mathcal{D}$. For an input with image $x_v$, let $h_{\text{base}}(x_v), h_{\text{ft}}(x_v) \in \mathbb{R}^{d}$ be the backbone hidden state at a given (layer, response-position) cell. We decompose the induced shift into an input-agnostic and an input-dependent term:

\vspace{-0.5em} 
\begin{align}
\label{eq:decomposition}
    h_{\text{ft}}(x_v) &= h_{\text{base}}(x_v) + \Delta h(x_v), \nonumber \\
    \Delta h(x_v) &= \mu + \nu(x_v),
\end{align}
with $\mu \in \mathbb{R}^{d}$ constant across inputs and $\mathbb{E}_{x_v}[\nu(x_v)] = \mathbf{0}$. A large, stable $\mu$ is what we call a \domainprior{}.

\subsection{Why a Static Component Dominates}
Narrow-domain targets are low-entropy and concentrated on a small sub-vocabulary, so moving $\mu$ along the dominant target direction reduces the loss for \emph{all} samples at once; refining $\nu(x_v)$ would instead have to fit the much more variable visual manifold. The asymmetry biases optimisation toward $\mu$ and starves the signal that would otherwise sharpen $\nu(x_v)$.

\begin{proposition}[Bias toward a static component; informal]
\label{prop:convergence}
Under gradient flow on a low-entropy narrow-domain objective, the update along $\mu$ dominates that on $\nu(x_v)$, so the shift concentrates into a persistent direction $\bar{\delta}$.
\end{proposition}

We treat Proposition~\ref{prop:convergence} as motivation, not a result. It predicts three empirically testable properties: the trace should be (i) large, (ii) stable across inputs, and (iii) still present when the image carries no information. Section~\ref{subsec:persistent-domain-prior} tests them.

\subsection{\ourmethod{}: Isolating the Trace}
\label{subsec:actdiff-extraction}
For a probe set $\mathcal{P}$ we compute the per-cell mean activation difference
\begin{equation}
\label{eq:actdiff}
\widehat{\delta} = \mathbb{E}_{x \sim \mathcal{P}}\big[\, h_{\text{ft}}(x) - h_{\text{base}}(x) \,\big].
\end{equation}
To pick out $\mu$ specifically, we additionally drive both models with degraded images---shuffled, text-only, blank, and Gaussian noise $x_{\text{noise}}\!\sim\!\mathcal{N}(0,I)$. Such inputs carry no coherent visual structure, so the $\nu$ term contributes negligibly in expectation and $\widehat{\delta}$ becomes an estimate of $\mu$. We select the dominant cell by normalised difference norm and use it as the anchor.

\subsection{Inference-Time Causal Ablation}
\label{subsec:actdiff-causal-restoration}
Because the trace is a concrete direction we can test its causal role without retraining, by subtracting a scaled copy at the anchor during decoding:
\begin{equation}
\label{eq:actdiff-causal-ablation}
\tilde{h}_{\text{ft}} = h_{\text{ft}} - \alpha\,\bar{\delta}.
\end{equation}
We sweep $\alpha$ and compare three interventions at matched magnitude: the matched-domain trace, a random direction, and the mean trace of the other domains. The two controls test whether any vector of comparable norm would produce the same recovery. We score the intervened model on a standard generic-VLM benchmark (the recovery axis) and on domain evaluation (the specialisation-cost axis).

\section{Experiments}

\subsection{Experimental Setup}
\label{subsec:experimental-setup}

The primary VLM is Qwen2.5-VL-3B-Instruct~\citep{bai2025qwen25vl}; cross-family checks additionally use Qwen2-VL-2B~\citep{wang2024qwen2vl}, LLaVA-NeXT-Llama3-8B~\citep{li2024llavanext_strong}, and Llama-3.2-V-11B-Vision-Instruct~\citep{meta2024llama32vision} with their AdaptLLM~\citep{cheng2024adaptllm} domain-adapted checkpoints on \textsc{Biomed}, \textsc{Food}, and \textsc{Remote~Sensing}. Diagnostic and behavioural probes draw $256$-sample subsets from an AdaptLLM-internal \emph{domain-eval} pool ($2{,}000$ per domain, held out from training) and a $2{,}000$-sample \emph{generic-eval} pool of VisualWebInstruct samples shared across domains. Causal restoration is scored on MMStar Overall~\citep{chen2024mmstar} ($N{=}600$), the benchmark on which the AdaptLLM-Biomed collapse is most pronounced (see Limitations). Mitigation training uses LoRA over $20{,}000$ AdaptLLM samples for $375$ steps with defaults $\lambda_\mu{=}1.0$, $\lambda_{\text{KL}}{=}0.5$ and anchor $\ell^\star{=}21$. Full hyperparameter and manifest-construction details are in Appendix~\ref{app:experimental-details}; total wall-clock in Appendix~\ref{app:overhead}.

\subsection{The Trace Exists and Is Input-Agnostic}
\label{subsec:persistent-domain-prior}

\begin{figure}[!t]
  \centering
  \begin{subfigure}[t]{0.95\linewidth}
    \centering
    \includegraphics[width=0.90\linewidth]{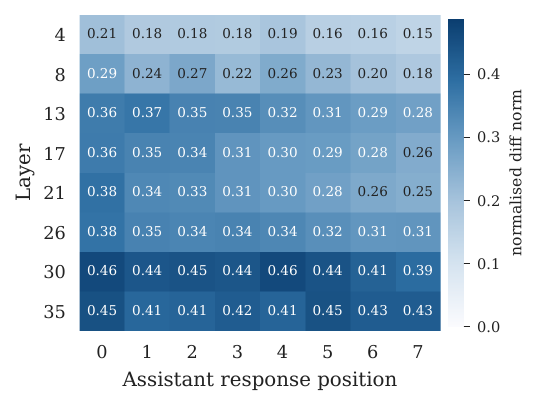}
    \caption{\textsc{Remote Sensing}}
    \label{fig:persistent-prior-rs}
  \end{subfigure}

  \vspace{0.45em}

  \begin{subfigure}[t]{0.95\linewidth}
    \centering
    \includegraphics[width=0.90\linewidth]{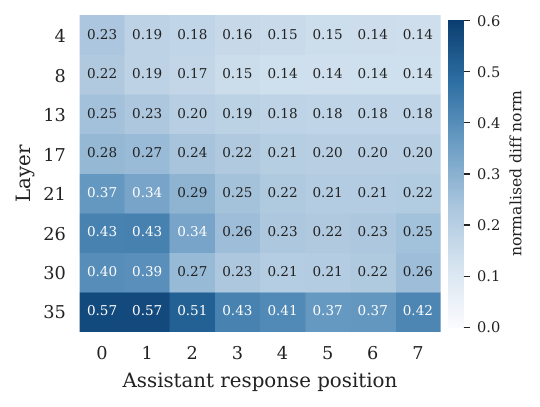}
    \caption{\textsc{Food}}
    \label{fig:persistent-prior-food}
  \end{subfigure}

  \vspace{0.45em}

  \begin{subfigure}[t]{0.95\linewidth}
    \centering
    \includegraphics[width=0.90\linewidth]{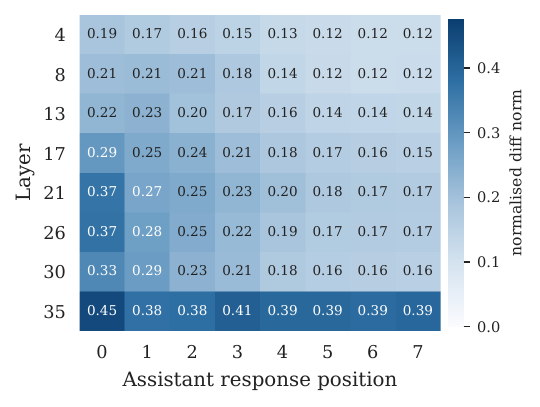}
    \caption{\textsc{Biomed}}
    \label{fig:persistent-prior-biomed}
  \end{subfigure}

  \caption{\textbf{Per-cell normalised activation-difference $\lVert\widehat{\delta}_{\ell,p}\rVert$ between AdaptLLM-finetuned and base Qwen2.5-VL-3B-Instruct on standard image--text inputs, for each of the three domains.} The offset is non-diffuse and concentrates at later layers and the earliest assistant response positions.}
  \label{fig:persistent-prior-heatmaps}
\end{figure}

\textbf{Narrow finetuning installs a compact internal residue that does not require an image to express itself.} Activation-diffing each base \vlm{} against its AdaptLLM finetuned counterpart on standard image--text inputs yields a structured, non-diffuse offset that concentrates at later language layers and the earliest assistant response positions. The strongest cells reach normalised difference norms of $0.586$ for \textsc{Food} (layer $35$, position $1$), $0.463$ for \textsc{Remote Sensing} (layer $30$, position $4$), and $0.452$ for \textsc{Biomed} (layer $35$, position $0$); see Figure~\ref{fig:persistent-prior-heatmaps}.

\begin{figure}[!tbp]
  \centering
  \includegraphics[width=\linewidth]{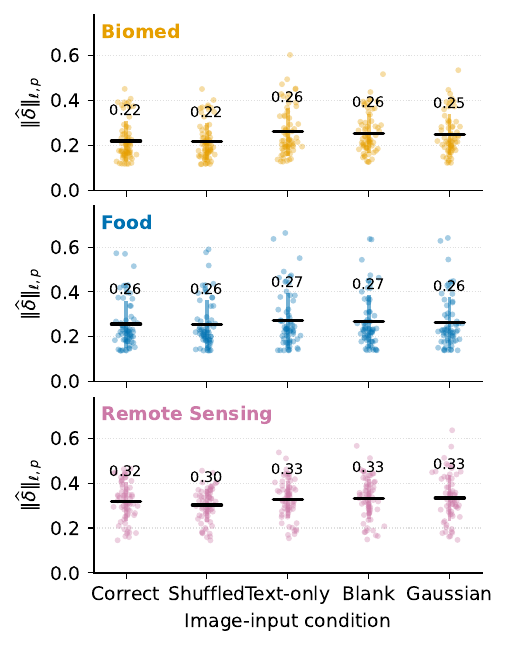}
  \caption{\textbf{Trace strength under progressively degraded image inputs.} The normalised difference distribution remains stable across correct, shuffled, text-only, blank, and Gaussian-noise inputs, showing that the dominant trace does not require image content.}
  \label{fig:persistent-prior-controls}
\end{figure}

The same extraction repeated under progressively degraded image inputs---shuffled, text-only, blank, and Gaussian noise---returns essentially the same offset. Averaged across the three domains the mean normalised difference norm is $0.265$ for correct image--text pairs, $0.259$ under shuffled images, $0.287$ under text-only prompts, $0.285$ under blank images, and $0.282$ under Gaussian noise (Figure~\ref{fig:persistent-prior-controls}). A trace that survives a blank image is by construction not an artefact of cross-modal alignment: it is the input-agnostic $\mu$ of Section~\ref{sec:method}.

This establishes the empirical precondition for the rest of the paper. The finetuned \vlm{} carries a stable domain-leaning default state that is not anchored to the actual image; whether that default is read out behaviourally as image-blindness, and whether removing it restores generic grounding, are taken up in Sections~\ref{subsec:illusion-of-expertise} and~\ref{subsec:causal-restoration}.

\subsection{Behaviourally, the Finetuned Model Looks Past the Image}
\label{subsec:illusion-of-expertise}

\textbf{If a finetuned VLM is leaning on an input-agnostic default rather than the image, destroying the image should hurt it less than it hurts the base.} For each (instruction, image) example we measure the per-condition negative log-likelihood (NLL) of the answer under five image variants---correct, same-domain counterfactual, cross-domain counterfactual, blank, and Gaussian noise---and compute the sensitivity as the NLL increase relative to the correct image. We aggregate over three seeds and report on the generic-eval and domain-eval pools defined in Section~\ref{subsec:experimental-setup}.

\begin{figure}[!t]
  \centering
  \begin{subfigure}[t]{\linewidth}
    \centering
    \includegraphics[width=0.86\linewidth]{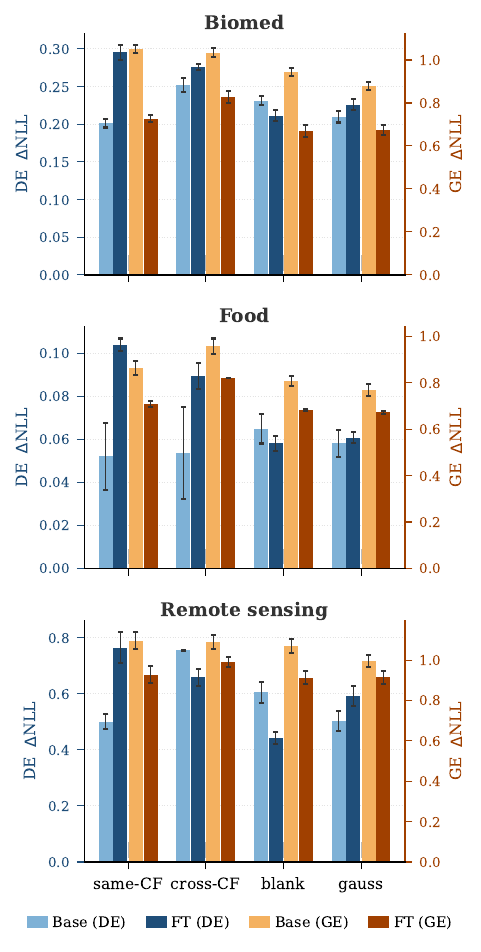}
    \caption{Per-condition image-corruption sensitivity ($\Delta\mathrm{NLL}$ vs.\ correct image), base vs.\ finetuned, across three domains and two evaluation pools. Error bars are $\pm 1$ s.d.\ over three seeds.}
    \label{fig:illusion-bars}
  \end{subfigure}

  \vspace{0.25em}

  \begin{subfigure}[t]{0.99\linewidth}
    \centering
    \includegraphics[width=0.95\linewidth]{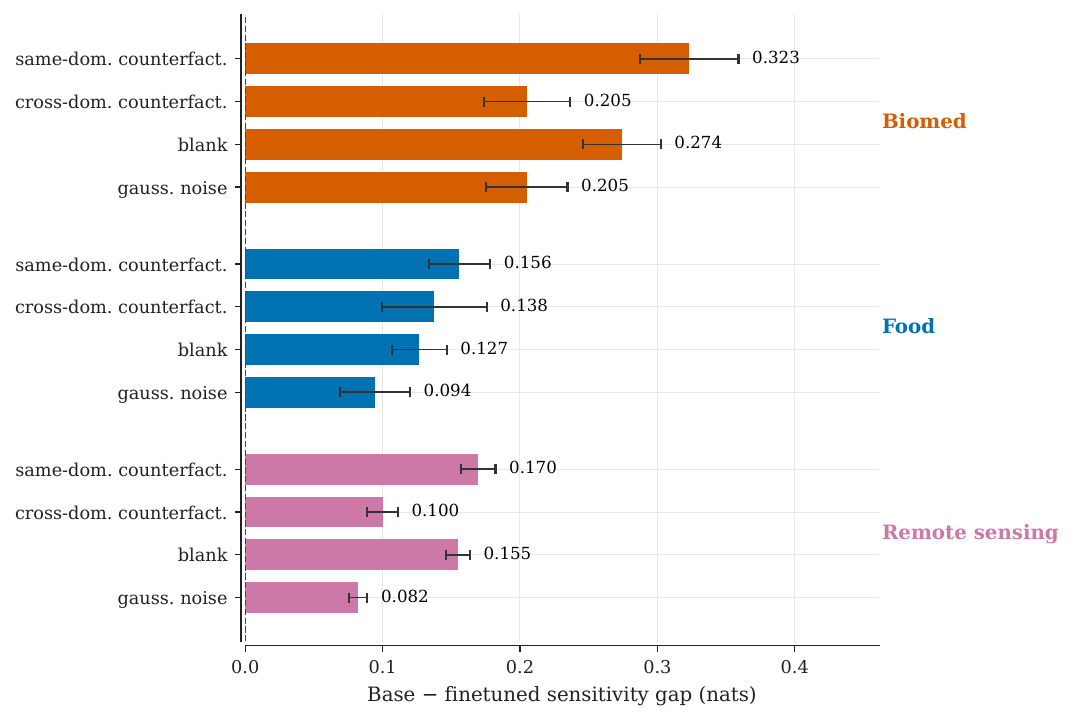}
    \caption{Base $-$ finetuned sensitivity gap under generic evaluation. Numeric labels are 3-seed means.}
    \label{fig:illusion-gap}
  \end{subfigure}

  \caption{\textbf{Finetuning reduces sensitivity to image corruption under OOD prompts.} On generic evaluation, the finetuned model is consistently less affected than the base across all corruptions and domains; the strongest gap is on \textsc{Biomed} under same-domain counterfactual images. The effect weakens under domain-matched evaluation, indicating OOD-specific rather than universal blindness.}
  \label{fig:illusion-of-expertise}
\end{figure}

The base--finetuned sensitivity gap is positive in every (domain, condition) cell under generic evaluation, in every seed (Figure~\ref{fig:illusion-bars}, right column; Figure~\ref{fig:illusion-gap}). On \textsc{Biomed} under same-domain counterfactual images the gap is $0.323 \pm 0.036$ nats (base $\Delta\mathrm{NLL} = 1.051 \pm 0.024$, finetuned $0.728 \pm 0.018$; per-seed gaps $0.293/0.315/0.363$). Averaged over the twelve generic-eval cells it is $0.169$ nats, and the smallest cell---\textsc{Remote Sensing} under Gaussian noise---is still $0.082 \pm 0.007$. The finetuned model has a default answer it can produce without the image, so corruption costs it less in expected NLL.





The same gap is \emph{not} universal under domain-eval: in $8$ of $24$ cells the finetuned model is in fact more sensitive than the base, most strongly on \textsc{Remote Sensing} same-domain counterfactual ($-0.265 \pm 0.035$). This is consistent with an overshadowing-default reading: when the prompt is itself domain-typical, the finetuned model engages with the image; when the prompt is out-of-distribution, it falls back to a domain-leaning answer. The illusion of expertise is therefore an out-of-distribution phenomenon, not a universal visual blindness.

As a behavioural footprint on a standard benchmark, base Qwen2.5-VL-3B-Instruct scores $0.550$ on MMStar; after AdaptLLM finetuning it scores $0.347$ on \textsc{Biomed} ($-37\%$ relative), $0.525$ on \textsc{Food} ($-4.5\%$), and $0.542$ on \textsc{Remote Sensing} ($-1.5\%$). The benchmark drop tracks the qualitative strength of the prior: the most aggressive medical finetune produces both the largest trace and the largest collapse. The causal link is the subject of Section~\ref{subsec:causal-restoration}. Behavioural NLL numbers in this section are on Qwen2.5-VL-3B-Instruct; running the same probe on LLaVA-NeXT-Llama3-8B and Llama-3.2-V-11B with the released finetunes (Appendix~\ref{sec:app-cross-family-exp11}) shows near-zero gaps. Rather than limiting our claim, this uncovers an architectural decoupling: the underlying activation-space trace universally recurs across all four families (Appendix~\ref{app:cross-model-robustness}), but its behavioural readout into image-blindness remains strictly architecture-gated.

\FloatBarrier
\subsection{The Trace Is Causally Implicated, Not a Correlate}
\label{subsec:causal-restoration}
\begin{figure*}[!h]
  \centering
  \includegraphics[width=0.84\linewidth]{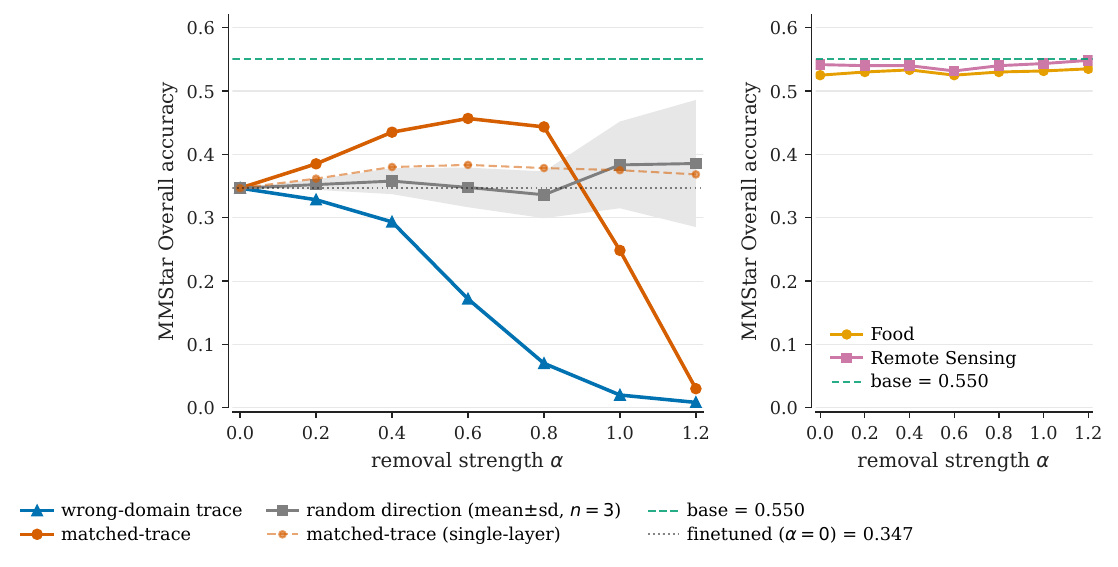}
  \caption{\textbf{Multi-layer trace ablation recovers part of the \textsc{Biomed} MMStar collapse with a peak-and-collapse dose response.} On \textsc{Biomed}, matched-trace removal peaks at $\alpha{=}0.6$, recovering $54.1\%$ of the $-20.3$-point collapse before dropping sharply for larger $\alpha$. The single-layer trace recovers much less, and the wrong-domain and random controls do not reproduce the same shape. On \textsc{Food} and \textsc{Remote Sensing}, where MMStar is not substantially affected by finetuning, the curves remain flat. $N{=}600$ MMStar items per configuration.}
  \label{fig:causal-restoration}
\end{figure*}
\textbf{Subtracting the extracted trace at inference recovers a substantial fraction of the benchmark collapse on the domain where the collapse is measurable, with a peak-and-collapse dose response that two structured controls do not produce.} We apply the ablation of Eq.~\eqref{eq:actdiff-causal-ablation} with $\alpha \in \{0.0, 0.2, 0.4, 0.6, 0.8, 1.0, 1.2\}$ and score the intervened model on MMStar ($N{=}600$). Two controls share the same anchor scheme and matched magnitude: a random direction per seed, and the mean trace of the other two AdaptLLM domains. We report two variants: single-layer ablation at the global best cell ($L{=}35,P{=}0$) and multi-layer ablation that hooks each of the eight stored layers $\{4,8,13,17,21,26,30,35\}$ with that layer's per-layer best-position delta. The trace itself is read from the input-agnostic source of Section~\ref{subsec:persistent-domain-prior} (gaussian-noise probe).

On \textsc{Biomed} (Figure~\ref{fig:causal-restoration}, left) the base scores $0.550$ and the finetune scores $0.347$ at $\alpha{=}0$, a $-37\%$ relative collapse worth $20.3$ accuracy points. Single-layer ablation recovers up to $0.383$ at $\alpha{=}0.6$, $18\%$ of the gap; multi-layer ablation recovers to $\mathbf{0.457}$ at $\alpha{=}0.6$, $\mathbf{54\%}$ of the gap ($3.0\times$ the single-layer recovery), then \emph{collapses catastrophically} to $0.248$ at $\alpha{=}1.0$ and $0.030$ at $\alpha{=}1.2$. The sharp peak-and-then-collapse shape is the qualitative signature of a specific causal direction: a small subtraction restores capability, an excessive one breaks the residual-stream feature the rest of the model expects.

The two controls do not reproduce this shape. The wrong-domain trace drives the model further below baseline at every $\alpha>0$ and reaches $0.008$ at $\alpha{=}1.2$. The random direction is more delicate: at the single seed plotted in Figure~\ref{fig:causal-restoration} it appears to rise to $0.438$ at $\alpha{=}1.0$ and $0.467$ at $\alpha{=}1.2$, but the 3-seed multi-seed retest (Appendix~\ref{app:multiseed-random-control}) shows this is a magnitude artefact: the means at $\alpha \in \{1.0, 1.2\}$ are $0.383 \pm 0.067$ and $0.386 \pm 0.103$, both within $\sim 1\sigma$ of the finetuned baseline $0.347$, with per-seed spreads of $10$--$14$ accuracy points. On no seed does the random direction reproduce the peak-and-collapse shape.


The same multi-layer ablation extends to two additional VLM families on the AdaptLLM-released biomed finetunes. LLaVA-NeXT-Llama3-8B recovers $+46.1\%$ of the collapse and peaks at the same $\alpha{=}0.6$ as Qwen; Llama-3.2-V-11B (Mllama) recovers only $+2.0\%$. We attribute this architectural split to fusion type: Qwen and LLaVA-NeXT inject vision tokens directly into the shared residual stream, so a static $\mu$ there can causally drive the collapse and be ablated. In contrast, Llama-3.2-V uses gated cross-attention, placing the shift outside the primary visual pathway, making the behavioral readout negligible. The split tracks the behavioural decomposition (Appendix~A.6): the families with a measurable image-grounding-loss component are recoverable, the family without it is not. Appendix~A.7 gives the per-family dose-response curves and layer choice.


On \textsc{Food} and \textsc{Remote Sensing} (Figure~\ref{fig:causal-restoration}, right) the finetune scores $0.525$ and $0.542$ at $\alpha{=}0$---only $-4.5\%$ and $-1.5\%$ relative to base---and the matched-trace curve is flat in $[0.51, 0.55]$ across the sweep. We read this as a null result by construction: the AdaptLLM finetunes on these domains do not measurably damage MMStar in the first place, so MMStar is the wrong axis on which to test causal recovery. The same domains do exhibit the trace (Section~\ref{subsec:persistent-domain-prior}) and the illusion-of-expertise gap on generic-eval NLL (Section~\ref{subsec:illusion-of-expertise}, $0.094$--$0.156$ nats).


\paragraph{Cross-benchmark Causal Recovery.} 
To confirm that the recovered capability is not specific to MMStar's item distribution, we extend the matched-trace multi-layer ablation to three additional standard visual grounding benchmarks on \textsc{Biomed} (Table~\ref{tab:four-benchmarks}): MMBench-EN \cite{Liu2023MMBenchIY}, RealWorldQA \cite{xai2024realworldqa}, and HallusionBench \cite{DBLP:conf/cvpr/GuanLWXLL0CHYM024}. The intervention reproduces the identical peak-and-collapse signature across all four axes. Notably, matched-trace ablation recovers $77\%$ of the performance collapse on RealWorldQA and $71\%$ on HallusionBench. This robust generalisation confirms that suppressing the input-agnostic prior broadly restores visual grounding.
\begin{table}[!htbp]
\centering\small
\setlength{\tabcolsep}{0pt} 
\caption{\textbf{Cross-benchmark causal recovery on \textsc{Biomed}.} The matched-trace ablation recovers substantial performance, reproducing the peak-and-collapse signature. Peak is at $\alpha{=}0.6$ except for HallusionBench ($\alpha{=}0.4$).}
\label{tab:four-benchmarks}
\begin{tabular*}{\columnwidth}{@{\extracolsep{\fill}}lcccc}
\toprule
Benchmark & Base & FT ($\alpha{=}0$) & Peak & Recovered \\
\midrule
MMStar & $0.555$ & $0.363$ & $0.457$ & $54\%$ \\
MMBench-EN & $0.800$ & $0.578$ & $0.707$ & $58\%$ \\
RealWorldQA & $0.658$ & $0.323$ & $0.580$ & $77\%$ \\
HallusionBench & $62.3$ & $28.2$ & $52.3$ & $71\%$ \\
\bottomrule
\end{tabular*}
\end{table}

\FloatBarrier

\subsection{The Prior Forms Early and Can Be Removed at Training Time}
\label{subsec:formation-mitigation}

\textbf{The persistent direction emerges within the first hundreds of optimisation steps, naive data mixing barely touches it, and a single first-moment regulariser shrinks the diagnostic trace by an order of magnitude while improving generic NLL.}

\begin{figure}[!t]
  \centering
  \includegraphics[width=\linewidth]{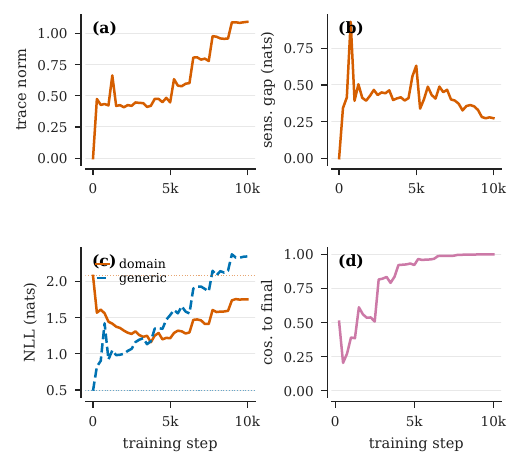}
  \caption{\textbf{The prior locks in early and keeps growing after behaviour saturates (\textsc{Biomed}, $10{,}000$ steps).} (a)~Best-cell norm rises from $0.47$ to $1.09$. (b)~Generic-eval sensitivity gap saturates early. (c)~Domain-target NLL drops while generic-target NLL rises. (d)~Cosine to the final anchor approaches $1.0$.}
  \label{fig:formation-dynamics}
\end{figure}

\paragraph{Formation dynamics.}
We re-finetune \textsc{Biomed} from the same base for $10{,}000$ steps and save adapter checkpoints every $250$ steps. At each checkpoint we re-extract the activation-difference grid against base and measure the best-cell normalised norm, its cosine to the final anchor direction, and the generic-eval image-sensitivity gap (Figure~\ref{fig:formation-dynamics}). The picture is two-phase: the \emph{direction} locks in immediately (cosine to final anchor reaches $0.51$ by step $250$, $0.92$ by step $5{,}000$, $0.999$ by step $9{,}000$) and the behavioural gap saturates within the first $\sim 1{,}000$ steps; the \emph{magnitude} keeps growing afterwards, with the anchor norm rising from $0.42$ at step $1{,}000$ to $1.09$ at step $10{,}000$ ($2.6\times$) long after behavioural collapse is essentially complete. The prior is installed early as a direction and merely sharpened by continued training, consistent with the gradient-flow account in Appendix~\ref{app:proofs}.

\begin{figure}[!t]
  \centering
  \includegraphics[width=\linewidth]{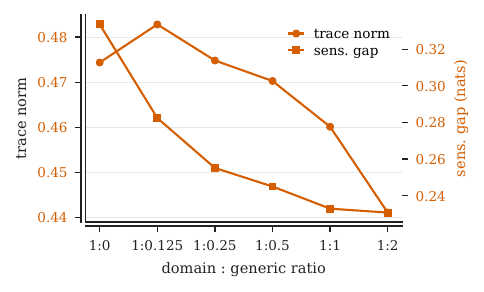}
  \caption{\textbf{Generic mixing is real but modest (\textsc{Biomed}, $20{,}000$ samples).} (a)~Best-cell norm changes little at $1{:}1$. (b)~Generic-eval sensitivity gap drops from $0.318$ to $0.246$ at $1{:}1$ ($-20\%$).}
  \label{fig:generic-mixing}
\end{figure}

\paragraph{Data-mixing baseline.}
The most obvious lever is to dilute the domain corpus with generic visual-instruction data. We finetune at six mixing ratios under a fixed $20{,}000$-sample budget and re-run the full diagnostic. The pure-domain run yields a generic-eval sensitivity gap of $0.318$ and a best-cell trace norm of $0.471$; a $1{:}1$ mixture reduces these to $0.255$ ($-20\%$) and $0.446$ ($-5\%$). Generic mixing is real but modest: even at $1{:}1$ the trace is essentially intact at the anchor cell (Figure~\ref{fig:generic-mixing}). The bar that the rest of this section has to clear.

\paragraph{Method-aligned regularisers.}
Section~\ref{subsec:persistent-domain-prior} and Appendix~\ref{app:proofs} both point to an input-agnostic mean shift $\mu$ as the source of grounding collapse. We propose two regularisers added to the standard cross-entropy loss at training time:
\begin{enumerate}\setlength\itemsep{0pt}
  \item \textbf{B1: $\mu$-shift.} At anchor layer $\ell^\star$ (early-layer per-cell peak from the diagnostic grid; $\ell^\star{=}21$ for all three domains), penalise the squared error between the batch-mean hidden states of the finetuned and base models: $\mathcal{L}_\mu = \lambda_\mu \cdot \mathrm{MSE}\!\big(\mathbb{E}_x[h_{\text{ft}}(x; \ell^\star)],\; \mathbb{E}_x[h_{\text{base}}(x; \ell^\star)]\big)$. Directly opposes Proposition~\ref{prop:convergence}.
  \item \textbf{B2: counterfactual KL.} For each batch, additionally forward the same prompts with the image replaced by a blank, and add $\mathcal{L}_{\text{KL}} = \lambda_{\text{KL}} \,\mathrm{KL}\!\big(P_{\text{base}}(\cdot \mid \text{text}, \varnothing) \,\|\, P_{\text{ft}}(\cdot \mid \text{text}, \varnothing)\big)$. Forces the finetuned model to fall back to base behaviour on uninformative images.
\end{enumerate}
Defaults $\lambda_\mu{=}1.0,\,\lambda_{\text{KL}}{=}0.5$ from a 3-point preliminary sweep on B1+B2. We compare against pure-domain and the $1{:}1$ generic-mixing baseline along three axes: anchor trace shrinkage, generic-eval NLL on correct images, and the generic-eval mean image-sensitivity gap (Table~\ref{tab:mitigation-summary}).

\begin{table}[!htbp]
\centering\small
\setlength{\tabcolsep}{5pt}
\caption{\textbf{Mitigation comparison across three AdaptLLM domains.} Anchor: $\ell^\star{=}21$, $p^\star{=}0$. Columns report anchor trace norm $\|\bar\delta\|$, correct-image NLL, and mean sensitivity gap; lower is better in all three.}
\label{tab:mitigation-summary}
\begin{tabular*}{\columnwidth}{@{\extracolsep{\fill}}lccc}
\toprule
Method & $\|\bar\delta\|$ & NLL & gap \\
\midrule
\multicolumn{4}{l}{\emph{\textsc{Biomed}}} \\
pure-domain               & $25.77$ & $0.872$ & $0.318$ \\
generic-mix $1{:}1$       & $24.91$ & $0.748$ & $0.255$ \\
B1                        & $0.71$  & $0.782$ & $0.241$ \\
B2                        & $25.18$ & $0.789$ & $0.355$ \\
B1$+$B2                   & $0.76$  & $0.732$ & $0.280$ \\
B2$'$-sameCF              & $19.35$ & $0.746$ & $0.314$ \\
B1$+$B2$'$-multi          & $0.79$  & $\mathbf{0.631}$ & $\mathbf{0.199}$ \\
\midrule
\multicolumn{4}{l}{\emph{\textsc{Food}}} \\
pure-domain               & $24.19$ & $0.728$ & $0.216$ \\
B1                        & $0.38$  & $0.656$ & $0.123$ \\
B1$+$B2                   & $0.41$  & $\mathbf{0.589}$ & $\mathbf{0.103}$ \\
\midrule
\multicolumn{4}{l}{\emph{\textsc{Remote~Sensing}}} \\
pure-domain               & $22.38$ & $0.743$ & $0.203$ \\
B1                        & $1.12$  & $0.664$ & $\mathbf{0.121}$ \\
B1$+$B2                   & $1.09$  & $\mathbf{0.621}$ & $0.165$ \\
\bottomrule
\end{tabular*}
\end{table}

\paragraph{Result.}
\textbf{B1 is the workhorse.} On the diagnostic axis it shrinks the anchor trace $\sim 34\times$ ($25.77 \to 0.71$); generic mixing moves it $\sim 3\%$ in comparison. \textbf{The joint B1+B2 gives the best capability:} generic-eval NLL drops from $0.872$ (pure) to $0.732$, better than generic mixing ($0.748$) and both ablations individually. Pulling the anchor toward base does not cost capability; it slightly improves it. \textbf{B2 alone fails informatively.} Its sensitivity gap rises rather than falls ($0.355$ vs.\ $0.318$ for pure), and the failure has a clean structural reading: B2 anchors only the conditional output on a single corruption fiber (the blank image), so with sufficient LoRA capacity the model can learn a ``blank-detector $\to$ base'' patch while preserving the input-agnostic $\mu$ on every other input mode. If the prior were input-specific, B2 should have worked and B1 would have been redundant; observing the opposite is a non-trivial empirical commitment to the input-agnostic $\mu$ of Proposition~\ref{prop:convergence}, and is formalised as a structural impossibility result in Appendix Proposition~\ref{prop:conditional-patch}. A $\lambda_\mu$ sensitivity sweep ($\lambda_\mu \in \{0.1,1.0,5.0\}$, $40$ steps, $2{,}000$-sample cap) shows monotonic trace shrinkage ($11.27 \to 2.53 \to 1.41$) with essentially flat $\mathcal{L}_{\text{CE}}$, so the regulariser tolerates more aggressive weighting if needed.


\paragraph{Comparison with SOTA Mitigations.}
We further benchmark B1 against representative weight-space regularization methods, including L2-SP \cite{DBLP:conf/icml/LiGD18}, L1-SP \cite{DBLP:conf/icml/LiGD18}, and EWC \cite{doi:10.1073/pnas.1611835114}, as well as a faithful single-domain adaptation of SMoLoRA's \cite{DBLP:conf/iccv/WangCWLSW25} routing mechanism. Table~\ref{tab:sota-comparison} shows the comparison under a matched training pipeline. While weight-space penalties suffer from capability drift or fail to suppress the input-agnostic prior (growing the trace by $4\times$ to $36\times$), B1 is the only method that effectively shrinks the trace while preserving both domain NLL and generic capability.

\begin{table}[!htbp]
\centering\small
\setlength{\tabcolsep}{0pt}
\caption{\textbf{Comparison with mitigation baselines on \textsc{Biomed}.} B1 shrinks the trace without compromising domain competence or generic grounding.}
\label{tab:sota-comparison}
\begin{tabular*}{\columnwidth}{@{\extracolsep{\fill}}lccc}
\toprule
Method & $\|\bar\delta\| \downarrow$ & Domain NLL $\downarrow$ & MMStar $\uparrow$ \\
\midrule
Pure-domain & $85.7$ & $1.470$ & $0.555$ \\
SMoLoRA & $329.3$ & $1.483$ & $0.545$ \\
L2-SP & $466.9$ & $1.625$ & $0.548$ \\
L1-SP & $881.4$ & $1.733$ & $0.545$ \\
EWC & $3060.9$ & $1.709$ & $0.555$ \\
\textbf{B1 (ours)} & $\mathbf{40.1}$ & $\mathbf{1.480}$ & $\mathbf{0.560}$ \\
\bottomrule
\end{tabular*}
\end{table}

\paragraph{Cross-domain replication (\textsc{Food}, \textsc{Remote Sensing}).}
The same trainer with the same defaults ($\lambda_\mu{=}1.0$, $\lambda_{\text{KL}}{=}0.5$, anchor $\ell^\star{=}21$, $p^\star{=}0$, $375$ steps on $20{,}000$ AdaptLLM samples) was rerun on \textsc{Food} and \textsc{Remote~Sensing} (Table~\ref{tab:mitigation-summary}, lower blocks). The three load-bearing patterns from \textsc{Biomed} reproduce on both domains: \textbf{(i)~B1 collapses the trace by an order of magnitude or more}---$64\times$ on \textsc{Food} ($24.19{\to}0.38$), $20\times$ on \textsc{Remote~Sensing} ($22.38{\to}1.12$), against $36\times$ on \textsc{Biomed}. \textbf{(ii)~B1$+$B2 keeps the trace shrunk and further improves capability}---NLL drops to $0.589$ on \textsc{Food} ($-10\%$ vs.\ B1) and $0.621$ on \textsc{Remote~Sensing} ($-7\%$), matching the $-6\%$ B1$+$B2-over-B1 NLL gain on \textsc{Biomed}. \textbf{(iii)~B1$+$B2 is the best mitigation overall in two of three domains}, taking the best NLL row on \textsc{Food} and \textsc{Remote~Sensing}; on \textsc{Biomed} the marginalised B1$+$B2$'$-multi variant edges it out, as Corollary~\ref{cor:marginalised-implies-b1} predicts. The image-sensitivity gap likewise improves on both new domains ($0.123{\to}0.103$ on \textsc{Food}, $0.121{\to}0.165$ on \textsc{Remote~Sensing}; the small uptick on RS mirrors the analogous $0.241{\to}0.280$ Biomed pattern, well below the no-mitigation baseline of $0.203$). The method-aligned regulariser is not Biomed-specific.

\begin{figure}[!htbp]
  \centering
  \includegraphics[width=0.95\linewidth]{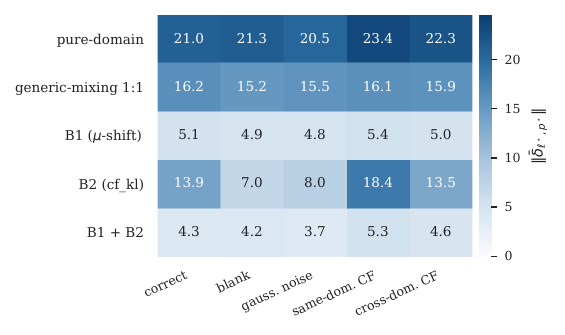}
  \caption{\textbf{Mitigation footprint by input condition.} At the anchor cell, pure-domain remains flat across fibers, B1 uniformly shrinks the trace, and B2 selectively suppresses blank/noise while leaving counterfactual fibers high, consistent with the conditional-patch mechanism of Proposition~\ref{prop:conditional-patch}.}
  \label{fig:fiber-localization}
\end{figure}

\paragraph{Structural test of B2.}
Two follow-ups discriminate the structural explanation from an engineering one. \emph{Fiber localisation} measures the anchor shift $\|\mathbb{E}_x[h_{\text{ft}}-h_{\text{base}}]\|_{\ell^\star,p^\star}$ separately on each of the five degraded-image input conditions. Pure-domain is flat at $\sim 21$ across fibers; B1 is uniformly flat at $\sim 5$; B2 is the only non-flat row---blank and Gaussian-noise fibers drop to $7.0$ and $8.0$ while same-domain CF rises to $18.4$ and cross-domain CF stays at $13.5$ (Figure~\ref{fig:fiber-localization}). The conditional-patch footprint is exactly what Proposition~\ref{prop:conditional-patch} predicts. \emph{We next} train two B2 variants under matched KL budget: B2$'$-sameCF replaces the blank fiber with same-domain counterfactual (the actual eval corruption), and B1+B2$'$-multi averages the KL over all four corruption fibers. B2$'$-sameCF still fails (trace $25.18 \to 19.35$, $\sim 23\%$ reduction, against $97\%$ for B1; sensitivity gap $0.314 \approx$ pure $0.318$): retargeting B2 to the eval fiber is not the fix. B1+B2$'$-multi matches B1 on trace shrinkage ($0.79$) and improves on it on both capability ($\mathrm{NLL}=0.631$ vs.\ B1's $0.782$) and image-grounding ($\mathrm{gap}=0.199$ vs.\ B1's $0.241$), exactly as Corollary~\ref{cor:marginalised-implies-b1} predicts (the marginalised KL is strictly stronger than B1's first-moment constraint).

\paragraph{Preservation of Benchmark Accuracy.}
To definitively close the evaluation-axis gap, we merged the LoRA adapters and evaluated the mitigation methods directly on standard VLM benchmarks (Table~\ref{tab:mitigation-accuracy}). The mitigations preserve base-level accuracy across MMStar, MMBench-EN, RealWorldQA, and MMMU\cite{DBLP:conf/cvpr/YueNZ0LZSJRSWYY24}. The training-time mitigation therefore carries no generic capability cost.

\begin{table}[!htbp]
\centering\small
\setlength{\tabcolsep}{0pt}
\caption{\textbf{Standard benchmark accuracy after mitigation (\textsc{Biomed}).} Merged adapters confirm that B1 and B1+B2 preserve base-level accuracy.}
\label{tab:mitigation-accuracy}
\begin{tabular*}{\columnwidth}{@{\extracolsep{\fill}}lccc}
\toprule
Benchmark & Base & B1 & B1$+$B2 \\
\midrule
MMStar & $0.555$ & $0.560$ & $0.557$ \\
MMBench-EN & $0.800$ & $0.797$ & $0.787$ \\
RealWorldQA & $0.658$ & $0.660$ & $0.658$ \\
MMMU & $0.422$ & $0.457$ & $0.455$ \\
\bottomrule
\end{tabular*}
\end{table}

\paragraph{Caveats.}
Three honest limitations. \textbf{(i)~Structural-test variants:} the three core methods (pure-domain, B1, B1$+$B2) are run on all three AdaptLLM domains, but the structural-test variants (B2$'$-sameCF and B1$+$B2$'$-multi) and B2 alone are only run on \textsc{Biomed}; their job is to isolate the conditional-patch mechanism, which we view as domain-agnostic given the matched trace behaviour we observe across domains. \textbf{(ii)~Anchor cell:} B1 collapses the trace at the chosen cell, but the global best-cell location of the grid migrates to layer~$35$ at convergence---a production-grade implementation would anchor on multiple cells jointly. \textbf{(iii)~Seed:} Table~\ref{tab:mitigation-summary} reports a single training seed per method. Appendix~\ref{app:multiseed-stability} retrains each method at two additional seeds and re-evaluates all three under the identical evaluation pipeline, finding $\sigma(\mathrm{NLL}) \leq 0.017$ and $\sigma(\mathrm{gap}_{\text{sens}}) \leq 0.019$ per method, roughly an order of magnitude smaller than the between-method differences; the ranking is preserved.


\FloatBarrier

\FloatBarrier

\section{Conclusion}

Narrow VLM finetuning installs an input-agnostic activation-space prior that the unembedding reads out as image-blindness. The prior has a clean diagnostic: it survives blank and Gaussian-noise images. It offers a causal handle at inference: subtracting it recovers $54\%$ of the \textsc{Biomed} MMStar collapse with a peak-and-collapse dose response absent in random controls. Finally, it enables a one-line training-time mitigation: a first-moment regulariser shrinks it $34\times$ and improves generic NLL beyond data-mixing baselines. The underlying trace universally recurs across four VLM families. While its behavioural translation into explicit image-blindness is architecture-gated, the core mechanistic failure remains fundamentally architecture-invariant. Beyond simply treating catastrophic forgetting as a weight-space problem, our findings demonstrate that safe multimodal adaptation requires explicit safeguards for the residual stream. As VLMs are increasingly deployed in high-stakes specialized domains, preserving their foundational visual integrity is just as critical as domain fluency. We hope the diagnostic and causal protocol give practitioners a handle for auditing and improving domain-finetuned VLMs.

\clearpage

\section*{Limitations}

\paragraph{Scope of empirical evidence.}
While the input-agnostic trace universally recurs at comparable magnitudes across all four VLM families tested, its behavioural readout into explicit image-blindness varies (prominent on Qwen2.5-VL-3B, attenuated on Llama-3.2-V-11B, and latent on LLaVA-NeXT-Llama3-8B). We interpret this as an architectural modulation of symptoms, not an absence of the underlying pathology; the cross-family causal ablation naturally inherits this split. Furthermore, the dose-response curve and the $54\%$ recovery figure are measured on MMStar Overall ($N{=}600$), chosen because the \textsc{Biomed} collapse is most pronounced there ($-37\%$ relative), yielding the cleanest signal. Extending this evaluation to other general-purpose benchmarks (MMBench, MMMU, SEED-Bench) and alternative narrow-finetuning recipes (e.g., LLaVA-Med-style two-stage or full-parameter SFT) remains valuable future work.

\paragraph{Scope of the mechanism and mitigation.}
That subtracting a single linear direction recovers $54\%$ of the \textsc{Biomed} collapse confirms the input-agnostic prior is a primary driver of the degradation. The remaining $46\%$ indicates it is not the sole cause; weight-level erasure, modality-bottleneck drift, and evaluation-distribution shift are not isolated by our pipeline and likely co-occur. The gradient-flow account is a motivation for the empirical predictions we test, not an asymptotic guarantee; only the conditional-patch result is stated and verified formally. On the mitigation side, B1 anchors the trace at one cell ($\ell^\star{=}21$); since the global best-cell location migrates to layer~35 at convergence, a production implementation should anchor jointly across layers. Finally, the structural-test variants (B2 alone, B2$'$-sameCF, B1$+$B2$'$-multi) are evaluated only on \textsc{Biomed}; cross-domain replication is left to future work.

\paragraph{Risk of misuse.}
\actdiff{} is a diagnostic and B1 is a first-moment regulariser; both inherit the generic risks of any model-editing technique (e.g.\ B1 could in principle mask other input-agnostic features such as safety priors by anchoring at a cell where they live), but we see no risk vector specific to our setting beyond those standard concerns.

\FloatBarrier

\section*{Acknowledgments}
This work was supported in part by the National Natural Science Foundation of China (NSFC) under Grant 62276283, in part by the China Meteorological Administration’s Science and Technology Project under Grant CMAJBGS202517, in part by the Guangdong-Hong Kong-Macao Greater Bay Area Meteorological Technology Collaborative Research Project under Grant GHMA2024Z04, in part by Fundamental Research Funds for the Central Universities, Sun Yat-sen University under Grant 23hytd006 and 23hytd006-2, in part by Guangdong Provincial High-Level Young Talent Program under Grant RL2024-151-2-11, and in part by the Key Development Project of the Artificial Intelligence Institute, Sun Yat-sen University under Grant 2025RGZN009.

\bibliography{bib/references}

\clearpage
\appendix
\section{Appendix}
\subsection{Mechanistic Motivation: Derivations}
\label{app:proofs}
\begin{strip}
\begin{tcolorbox}[
    enhanced, breakable,
    colback=gray!2, colframe=black!50,
    boxrule=0.4pt, arc=2pt,
    left=10pt, right=10pt, top=8pt, bottom=8pt,
]

These derivations support the compact account in Section~\ref{sec:method}. We present them as motivation: they explain \emph{why} narrow finetuning biases the model toward an input-agnostic component and how such a component can suppress generic generation.

\paragraph{Gradient-flow argument for Proposition~\ref{prop:convergence}.}
Decompose the finetuned hidden state as $h(x) = \mu + \nu(x)$ with $\mu$ input-agnostic and $\mathbb{E}_x[\nu(x)] = \mathbf{0}$, and model finetuning as gradient flow on the cross-entropy loss $L = -\log P(y^\ast \mid x)$ with unembedding rows $W_j$. The hidden-state gradient is $\nabla_h L = \sum_{j}\big(P(j \mid x) - \mathbb{I}_{j = y^\ast}\big) W_j$, giving
\begin{equation}
\frac{d\mu}{dt} = \mathbb{E}_{x \sim \mathcal{D}}\!\Big[W_{y^\ast} - \!\sum_j P(j \mid x) W_j\Big], \qquad
\frac{d\nu(x)}{dt} = W_{y^\ast} - \!\sum_j P(j \mid x) W_j .
\end{equation}
Under the narrow-domain assumption the targets $y^\ast$ concentrate on a sub-vocabulary, so $\mathbb{E}_{x \sim \mathcal{D}}[W_{y^\ast}]$ aligns with a dominant direction $W_{\mathcal{D}}$. Moving $\mu$ along $W_{\mathcal{D}}$ lowers the loss for all samples simultaneously, and the implicit bias of gradient descent drives $\lVert \mu(t) \rVert$ to grow. As $\lVert \mu \rVert$ grows, $P(y^\ast \mid x) \to 1$ and the error signal $e(x) = 1 - P(y^\ast \mid x)$ decays; since $\lVert d\nu(x)/dt \rVert \le e(x) \lVert W_{y^\ast} \rVert$, the dynamic component is starved before it can fit the visual manifold. Hence $\lim_{t \to \infty} \Delta h_t(x) \approx \mu_\infty \equiv \bar{\delta}$, a persistent input-agnostic prior.

\paragraph{A suppression bound.}
Let $h_{\text{ft}} = h_{\text{base}} + \bar{\delta}$ and $z = W_U h_{\text{ft}}$. For a generic concept $c$ and a domain token $k$, dropping the positive remaining vocabulary mass gives $P(c) < \sigma\!\big(-(z_k - z_c)\big)$. Substituting $z_k - z_c$ and writing $\gamma = (W_U^{(k)} - W_U^{(c)})^\top \bar{\delta} / \lVert \bar{\delta} \rVert$ and $C_{\text{base}} = (W_U^{(c)} - W_U^{(k)})^\top h_{\text{base}}$ yields
\begin{equation}
P(c) < \frac{1}{1 + \exp\!\big(\gamma\,\lVert \bar{\delta} \rVert - C_{\text{base}}\big)} .
\end{equation}
Because $\bar{\delta}$ is optimized to promote domain tokens, $\gamma > 0$; thus a larger input-agnostic shift can exponentially down-weight generic concepts at the unembedding bottleneck, consistent with pretrained knowledge being \emph{overshadowed} rather than erased. We treat this as a mechanistic rationale, not as a measured quantity.

\paragraph{Setup for the conditional-patch result.}
\label{app:cp-setup}
Fix a layer index $\ell^\star$. Let $(\mathcal{X}, \Sigma, \mathcal{D})$ be the input probability space and let $\mathcal{F}$ denote a class of bounded measurable maps $f : \mathcal{X} \to \mathbb{R}^d$ realisable by the parameter family $\{f_\theta : \theta \in \Theta\}$ at $\ell^\star$. Let $W_U \in \mathbb{R}^{V \times d}$ be a fixed unembedding matrix with $\mathrm{rank}(W_U) \leq d$ and define the conditional response
\begin{equation}
P_\theta(\cdot \mid x) \;:=\; \mathrm{softmax}\!\big(W_U f_\theta(x)\big), \qquad x \in \mathcal{X}.
\end{equation}
Let $\theta_b \in \Theta$ denote the base parameters, write $g_\theta(x) := f_\theta(x) - f_{\theta_b}(x)$ for the residual-stream shift induced by finetuning, and define the input-averaged shift $\mu_\theta := \mathbb{E}_{x \sim \mathcal{D}}\!\left[g_\theta(x)\right] \in \mathbb{R}^d$. Boundedness of $\mathcal{F}$ guarantees integrability of $g_\theta$.

\refstepcounter{lemma}\label{lem:softmax-kernel}
\paragraph{Lemma~\thelemma\ (Kernel of $\mathrm{softmax}\circ W_U$).}
For any $u, v \in \mathbb{R}^d$,
\begin{equation}
\mathrm{KL}\!\big(\mathrm{softmax}(W_U u) \,\big\|\, \mathrm{softmax}(W_U v)\big) \;=\; 0
\;\Longleftrightarrow\;
W_U(u - v) \in \mathrm{span}(\mathbf{1}_V).
\end{equation}

\emph{Proof.} ($\Leftarrow$) $\mathrm{softmax}(z + c \mathbf{1}_V) = \mathrm{softmax}(z)$ for any $c \in \mathbb{R}$ since the additive constant cancels in normalisation. ($\Rightarrow$) $\mathrm{KL}(p \,\|\, q) = 0$ with $p, q$ probability vectors implies $p = q$; the preimage of any point under $z \mapsto \mathrm{softmax}(z)$ on $\mathbb{R}^V$ is an affine line parallel to $\mathbf{1}_V$.

\refstepcounter{proposition}\label{prop:conditional-patch}
\paragraph{Proposition~\theproposition\ (Conditional-patch lemma for fiber-specific KL).}
Let $F \in \Sigma$ with $\alpha := \mathcal{D}(F) < 1$. Assume:
\begin{itemize}\setlength\itemsep{1pt}
  \item[\textbf{(A1)}] (\emph{constant-on-complement realisability}) For every $v \in \mathbb{R}^d$, the function $x \mapsto f_{\theta_b}(x) + \mathbf{1}_{\mathcal{X} \setminus F}(x)\,v$ belongs to $\mathcal{F}$.
\end{itemize}
Then for every $\mu^\star \in \mathbb{R}^d$ there exists $\theta^\star \in \Theta$ such that
\begin{align}
  \mathrm{KL}\!\big(P_{\theta_b}(\cdot \mid x) \,\big\|\, P_{\theta^\star}(\cdot \mid x)\big) &= 0 \qquad \text{for $\mathcal{D}$-a.e.\ } x \in F, \label{eq:p2-fiber} \\
  \mu_{\theta^\star} \;=\; \mathbb{E}_{x \sim \mathcal{D}}\!\big[g_{\theta^\star}(x)\big] &= (1 - \alpha)\,\mu^\star. \label{eq:p2-marginal}
\end{align}

\emph{Proof.} By (A1) choose $\theta^\star \in \Theta$ realising $f_{\theta^\star}(x) = f_{\theta_b}(x) + \mathbf{1}_{\mathcal{X} \setminus F}(x)\,\mu^\star$, so $g_{\theta^\star}(x) = \mathbf{1}_{\mathcal{X} \setminus F}(x)\,\mu^\star$.

For~\eqref{eq:p2-fiber}: on $x \in F$, $g_{\theta^\star}(x) = 0$, hence $W_U(f_{\theta^\star}(x) - f_{\theta_b}(x)) = 0 \in \mathrm{span}(\mathbf{1}_V)$, hence by Lemma~\ref{lem:softmax-kernel}, $\mathrm{KL} = 0$.

For~\eqref{eq:p2-marginal}: $\mathbb{E}_{x \sim \mathcal{D}}[\mathbf{1}_{\mathcal{X} \setminus F}(x)\,\mu^\star] = (1 - \mathcal{D}(F))\,\mu^\star = (1 - \alpha)\,\mu^\star$.

\refstepcounter{corollary}\label{cor:no-fiber-works}
\paragraph{Corollary~\thecorollary\ (No fiber-specific KL constraint forces $\mu_\theta = 0$).}
Under (A1), let
\begin{equation}
\mathcal{C}_F \;:=\; \big\{\theta \in \Theta \,:\, \mathrm{KL}\!\big(P_{\theta_b}(\cdot \mid x) \,\big\|\, P_\theta(\cdot \mid x)\big) = 0 \text{ for $\mathcal{D}$-a.e.\ } x \in F\big\}.
\end{equation}
For any $F \in \Sigma$ with $\mathcal{D}(F) < 1$,
\begin{equation}
\mathcal{C}_F \;\not\subseteq\; \big\{\theta \in \Theta \,:\, \mu_\theta = 0\big\}.
\end{equation}

\emph{Proof.} Apply Proposition~\ref{prop:conditional-patch} with $\mu^\star \neq 0$; the constructed $\theta^\star$ satisfies $\theta^\star \in \mathcal{C}_F$ but $\mu_{\theta^\star} = (1-\alpha)\mu^\star \neq 0$.

\refstepcounter{corollary}\label{cor:marginalised-implies-b1}
\paragraph{Corollary~\thecorollary\ (Full-support KL is strictly stronger than B1).}
Define the full-support KL constraint set and the B1 first-moment constraint set:
\begin{align}
\mathcal{C}^{\mathrm{full}} &\;:=\; \Big\{\theta : \mathbb{E}_{x \sim \mathcal{D}}\!\big[\mathrm{KL}\!\big(P_{\theta_b}(\cdot \mid x) \,\big\|\, P_\theta(\cdot \mid x)\big)\big] = 0\Big\}, \\
\mathcal{C}^{B1} &\;:=\; \big\{\theta : W_U \mu_\theta \in \mathrm{span}(\mathbf{1}_V)\big\}.
\end{align}
Then $\mathcal{C}^{\mathrm{full}} \subseteq \mathcal{C}^{B1}$, and the inclusion is strict whenever $\mathcal{F}$ contains a non-constant $g_\theta$ satisfying $W_U g_\theta(x) \in \mathrm{span}(\mathbf{1}_V)$ for $\mathcal{D}$-a.e.\ $x$ \emph{and} $W_U g_\theta(x) \neq 0$ on a positive-measure subset.

\emph{Proof.}
\emph{Inclusion.} Let $\theta \in \mathcal{C}^{\mathrm{full}}$. The integrand $\mathrm{KL}(P_{\theta_b}(\cdot \mid x) \,\|\, P_\theta(\cdot \mid x))$ is non-negative and $\mathcal{D}$-integrable; its expectation is zero, so the integrand vanishes $\mathcal{D}$-a.e. By Lemma~\ref{lem:softmax-kernel}, $W_U g_\theta(x) \in \mathrm{span}(\mathbf{1}_V)$ for $\mathcal{D}$-a.e.\ $x$. Writing $W_U g_\theta(x) = c(x) \mathbf{1}_V$ for a measurable scalar function $c : \mathcal{X} \to \mathbb{R}$ on the full-measure set, linearity of the Bochner expectation gives $W_U \mu_\theta = \mathbb{E}_x[W_U g_\theta(x)] = \mathbb{E}_x[c(x)]\,\mathbf{1}_V \in \mathrm{span}(\mathbf{1}_V)$, so $\theta \in \mathcal{C}^{B1}$.

\emph{Strictness.} Take any $\theta \in \mathcal{F}$ with the stated non-constant $g_\theta$. Then $\theta \in \mathcal{C}^{B1}$ but $W_U g_\theta(x) \not\equiv 0$ on a positive-measure set, so $\mathrm{KL}(P_{\theta_b}(\cdot|x) \,\|\, P_\theta(\cdot|x))$ need not vanish on that set, i.e.\ $\theta$ can fail $\mathcal{C}^{\mathrm{full}}$.

\paragraph{Remark on assumption (A1).}
(A1) is the only substantive assumption. It requires the model class to realise the function $x \mapsto f_{\theta_b}(x) + \mathbf{1}_{\mathcal{X} \setminus F}(x)\,v$ for an arbitrary direction $v \in \mathbb{R}^d$. In a neural-network setting, this decomposes into a soft indicator $\hat{\mathbf{1}}_{\mathcal{X} \setminus F}$ computable in the hidden state at or before $\ell^\star$ and one additional linear direction in the residual stream parameterised by $v$. The proposition is exact under the idealised hard indicator; for any $\epsilon > 0$, a soft indicator agreeing with $\mathbf{1}_F$ off a set of measure $\leq \epsilon$ yields a KL bounded by a constant multiple of $\epsilon$, by continuity of softmax-KL on $\mathbb{R}^d$ and boundedness of $\mathcal{F}$. When $F$ is defined by a low-dimensional input statistic computable from the post-encoder feature---for instance blank-image or Gaussian-noise inputs, both of which collapse the visual-encoder output norm---a rank-$1$ LoRA adapter on the language-model side suffices for (A1). Our structural test (Section~\ref{subsec:formation-mitigation}) empirically verifies that the LoRA adapter trained with $\mathcal{L}_{\text{KL}}$ on blank-image inputs realises the construction of Proposition~\ref{prop:conditional-patch} to within measurement precision at the anchor cell.

\paragraph{Interpretation tying back to Section~\ref{subsec:formation-mitigation}.}
Proposition~\ref{prop:conditional-patch} gives a formal counterpart to the per-fiber footprint of the structural test (Figure~\ref{fig:fiber-localization}). The construction $g_{\theta^\star}(x) = \mathbf{1}_{\mathcal{X} \setminus F}(x)\,\mu^\star$ is precisely the conditional-patch profile observed in the cf\_kl row of Figure~\ref{fig:fiber-localization}: $\lVert g_{\theta^\star} \rVert$ small on $F = \{$blank, Gaussian noise$\}$ and at the pre-finetuning level on $\mathcal{X} \setminus F = \{$correct, same-domain CF, cross-domain CF$\}$. Corollary~\ref{cor:no-fiber-works} elevates the empirical failure of B2 into a topological non-implication: under (A1), no fiber strictly smaller than $\mathcal{X}$ admits an alternative outcome---B2 cannot in principle remove an input-agnostic $\mu$. Corollary~\ref{cor:marginalised-implies-b1} clarifies the only remaining KL-style route: the full-support KL constraint $\mathcal{C}^{\mathrm{full}}$ is strictly tighter than the B1 first-moment constraint $\mathcal{C}^{B1}$; it implies B1 through its first moment, and exceeds B1 by forbidding any input-conditional softmax-equivalent shift---including the legitimate $\nu(x)$ component of Section~\ref{subsec:persistent-domain-prior}. The B1 ($\mu$-shift) regulariser therefore sits at the minimal sufficient point on the KL hierarchy: it is the unique first-moment projection that any input-agnostic-prior-removing KL constraint must imply, while being weak enough to coexist with task-relevant input-dependent variation $\nu(x)$.

\end{tcolorbox}
\end{strip}

\subsection{Cross-Family Robustness: Per-Model Trace Maps}
\label{app:cross-model-robustness}

To show that the persistent trace is not an artifact of a single checkpoint, we replicate the activation-difference extraction across a completed matched matrix of \textbf{four VLM families spanning 2B--11B parameters and three architectures} (\textsc{Qwen2-VL-2B}, \textsc{Qwen2.5-VL-3B}, \textsc{LLaVA-NeXT-Llama3-8B}, \textsc{Llama-3.2-Vision-11B}) crossed with the three AdaptLLM domains, i.e.\ a complete $4\times 3$ matrix of 12 matched base/finetuned pairs. Rather than collapsing each pair to a single scalar, we plot the full layer$\times$position normalized-difference heatmap for every cell, in the same style as the main-text persistent-trace figure, so the reader can inspect the trace geometry family by family.

\begin{figure*}[!tbp]
  \centering
  \includegraphics[width=0.95\linewidth]{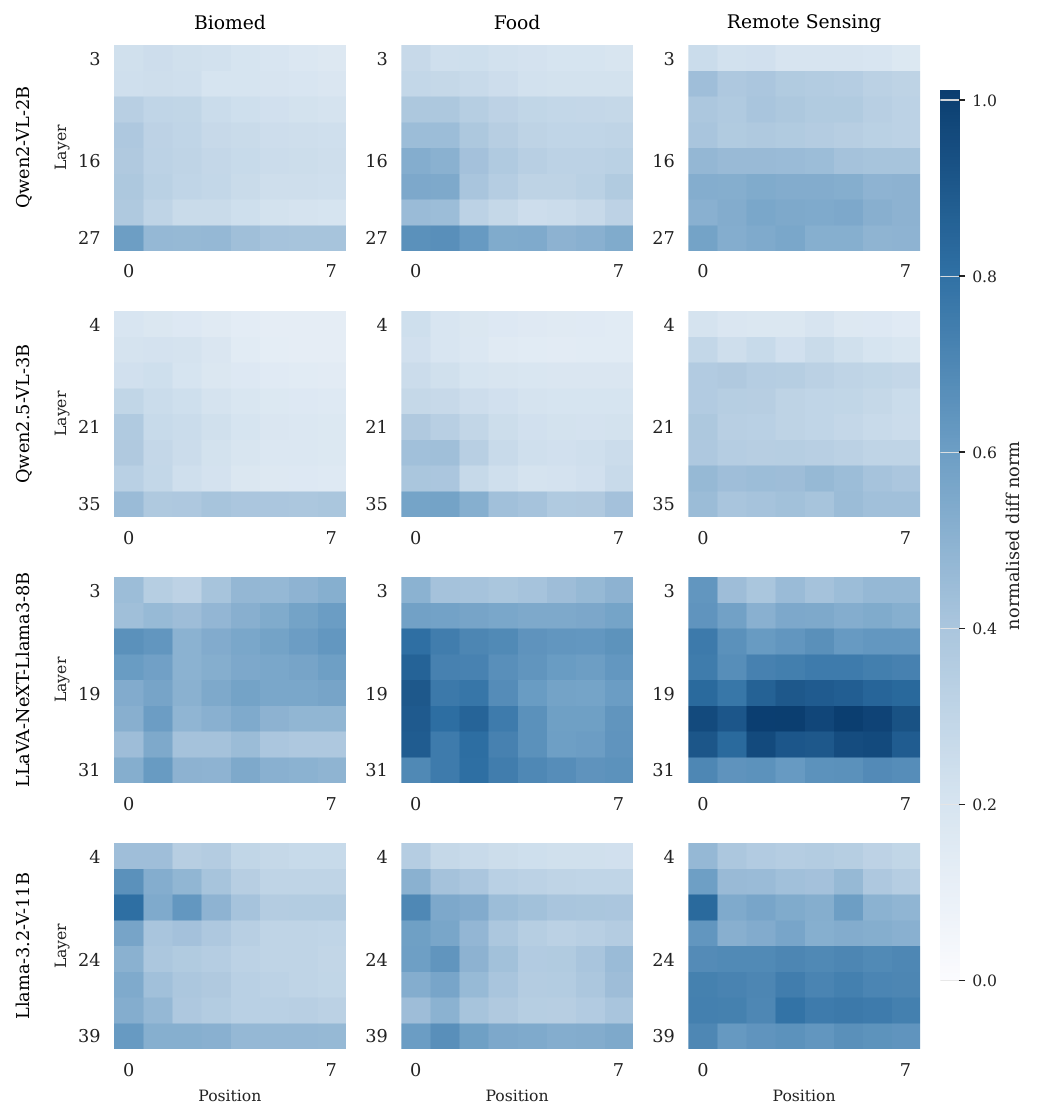}
  \caption{
    \textbf{Cross-family persistent trace, drawn family by family.} Full layer$\times$assistant-position normalized-difference-norm heatmaps for the complete $4\times3$ matched matrix (rows: VLM families ordered by scale; columns: AdaptLLM domains). A shared color scale makes cross-family magnitude differences directly comparable. All twelve cells show a structured, non-diffuse trace concentrated at early response positions; strength is largest for the \textsc{LLaVA-NeXT} and \textsc{Llama-3.2-Vision} families and smaller but still clear for the \textsc{Qwen} families. This supports a calibrated generality claim: the persistent domain prior recurs across families and scales, with heterogeneous magnitude and a shared compact early-position locus.
  }
  \label{fig:app-cross-model-grid}
\end{figure*}

Figure~\ref{fig:app-cross-model-grid} shows the result. Three observations hold across all twelve cells. First, every matched pair exhibits a clear non-diffuse trace, so the phenomenon is not confined to one family or scale. Second, the magnitude is heterogeneous but structured: the larger \textsc{LLaVA-NeXT-Llama3-8B} and \textsc{Llama-3.2-Vision-11B} models show the strongest traces (peak normalized-difference norm up to $1.01$ and $0.83$ respectively), while the \textsc{Qwen} families are consistently weaker but still clearly non-zero (peaks $0.45$--$0.66$). Third, the trace remains spatially compact: within each family the dominant cells concentrate at the earliest assistant response positions, with family-specific layer locations. We therefore read this matrix as evidence for \emph{cross-family recurrence with heterogeneous strength}, not strict numerical invariance.

\subsection{The Trace Carries Domain-Identifying Information}
\label{app:domain-identifying}

Sections~\ref{subsec:persistent-domain-prior}--\ref{subsec:illusion-of-expertise} show the trace exists and is input-agnostic. Here we check that it is \emph{domain-specific structure} rather than undifferentiated finetuning noise. We treat each (seed, domain) trace as a sample and perform leave-one-seed-out 3-way identification of the finetuning domain.

\begin{figure}[!tbp]
  \centering
  \begin{subfigure}[t]{0.38\linewidth}
    \centering
    \includegraphics[width=\linewidth]{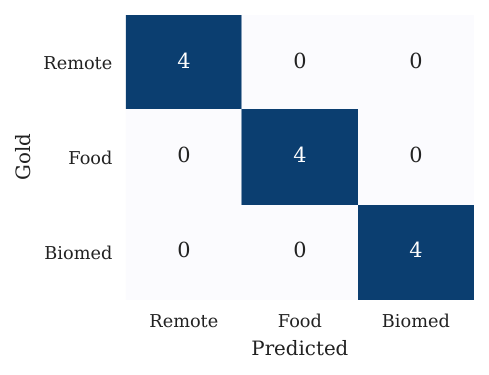}
    \caption{Ensemble domain identification}
    \label{fig:app-domain-id-conf}
  \end{subfigure}
  \hfill
  \begin{subfigure}[t]{0.58\linewidth}
    \centering
    \includegraphics[width=\linewidth]{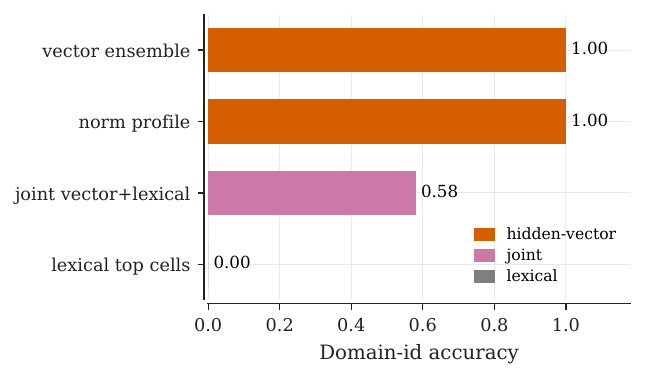}
    \caption{Readout method comparison}
    \label{fig:app-domain-id-method}
  \end{subfigure}
  \caption{
    \textbf{The trace is domain-identifying, geometrically rather than lexically.} \textbf{(a)} Leave-one-seed-out 3-way identification of the finetuning domain from the trace yields an exactly diagonal confusion matrix (ensemble accuracy $1.00$, $n=12$; three domains, four seeds). \textbf{(b)} Comparing readout methods: hidden-vector methods identify the finetuning domain at $1.00$, while a lexical top-cell token readout is at $0.00$ and a joint method at $0.58$. The discriminative signal lives in activation geometry, not in promoted tokens.
  }
  \label{fig:app-domain-id}
\end{figure}

In our setup---three AdaptLLM domains, four seeds---a nearest-centroid readout over the trace recovers the finetuning domain perfectly: the aggregate confusion matrix is exactly diagonal (Figure~\ref{fig:app-domain-id}a; ensemble accuracy $1.00$, $n=12$), and accuracy is at ceiling across all probed cells. Because the label space here is only three domains, we read this as strong evidence that the trace is cleanly domain-separable, not as a calibrated error rate. Further analysis sharpens the point by comparing \emph{how} the domain is read out (Figure~\ref{fig:app-domain-id}b): hidden-vector methods reach $1.00$ accuracy (vector ensemble and norm profile), whereas a lexical top-cell token readout reaches $0.00$ and a joint lexical+vector method only $0.58$. The information that identifies the finetuning domain is therefore carried \emph{geometrically} in the activation difference, not on the surface via promoted tokens---consistent with the partial, noisy logit-lens readouts noted in the main text. We include these as supporting evidence: they reinforce that the persistent prior is structured, domain-specific content.

\subsection{Compute and Wall-Clock Overhead}
\label{app:overhead}

Reviewers reasonably ask what the proposed regularisers and inference-time ablation cost in practice. We report both an asymptotic FLOPs-style count of the extra forwards each method requires and the measured wall-clock from our actual training and inference runs.

\paragraph{Theoretical accounting.}
Let one ``base step'' denote a single forward$+$backward through the finetuned LoRA-augmented model on one micro-batch. The proposed regularisers add unmodified base-model forwards (no gradient) and, for B2-family methods, additional finetuned-model forwards on a corrupted copy of the batch. All extra forwards reuse the same LoRA parameters; no new trainable parameters are introduced. Per micro-batch the additional cost is:
\begin{enumerate}\setlength\itemsep{1pt}
  \item \textbf{B1 ($\mu$-shift)}: $+1$ base forward on the real batch (no-grad), to evaluate $\mathbb{E}_x[h_{\text{ft}}(x) - h_{\text{base}}(x)]$. The regulariser term itself is $O(d)$ scalar arithmetic.
  \item \textbf{B2 (single-fiber KL)}: $+1$ base forward and $+1$ finetuned forward on the corrupted batch (the finetuned forward must be with grad through the KL term).
  \item \textbf{B1 $+$ B2}: union of the above. The B1 and B2 base forwards run on \emph{different} batches (real vs.\ corrupted), so they do not share compute.
  \item \textbf{B2$'$-multi}: identical to a single B2 step per micro-batch, with the corruption mode cycled round-robin across the four fibers (blank, Gaussian noise, same-domain counterfactual, cross-domain counterfactual). The expected gradient matches uniform-mixture averaging at $1/4$ the per-step cost; no compute overhead vs.\ B2-blank.
\end{enumerate}
Inference-time multi-layer ablation (Section~\ref{subsec:causal-restoration}, F2) adds $K$ forward hooks at $K$ chosen layers, each performing a single $d$-dimensional vector add on the residual stream. The arithmetic added per token is $O(K d)$, dominated by the $O(d^2)$ MLP and attention work at every layer. The forward FLOPs ceiling therefore grows by $O(K / L) \cdot (d / d^2) \approx 10^{-3}$ relative; the observable overhead in our PyTorch implementation comes from the per-hook Python callback latency rather than the arithmetic.

\paragraph{Measured wall-clock (Qwen2.5-VL-3B-Instruct, LoRA $r{=}64$, $\alpha{=}128$, $6000$ micro-steps $=$ $375$ outer steps with $16$-step gradient accumulation, single A100-64GB, bf16, gradient checkpointing on, batch size $1$).}
Table~\ref{tab:overhead-training} reports the training wall-clock reported by the Hugging Face Trainer for each mitigation method. We include both an absolute run time and the per-micro-batch step time, plus a ratio to B1 ($\mu$-shift) as the lightest method-aligned regulariser.

\begin{table*}[!tbp]
\centering\small
\setlength{\tabcolsep}{4.5pt}
\caption{
  \textbf{Training wall-clock per method on a single A100-64GB.} ``Extra fwd/step'' counts the auxiliary forward passes added by the regulariser on top of the standard finetuned forward$+$backward. Ratio is relative to B1.
}
\label{tab:overhead-training}
\begin{tabular}{lcccc}
\toprule
Method & extra fwd / step & total (h) & per-micro (s) & ratio vs.\ B1 \\
\midrule
B1: $\mu$-shift                    & $1$    & $1.69$ & $1.01$ & $1.00\times$ \\
B2: cf\_kl (blank)                 & $2$    & $3.47$ & $2.08$ & $2.05\times$ \\
B1 $+$ B2 (blank)                  & $3$    & $3.13$ & $1.88$ & $1.85\times$ \\
B2$'$-sameCF (retargeted)          & $2$    & $3.26$ & $1.95$ & $1.93\times$ \\
B1 $+$ B2$'$-multi (marginalised)  & $3$    & $3.36$ & $2.02$ & $1.99\times$ \\
\bottomrule
\end{tabular}
\end{table*}

The measured ratios track the extra-forward counts: methods with one extra base forward (B1) run at $\sim 1\times$, methods with two extra forwards on a corrupted batch (B2 family) run at $\sim 2\times$. The variability across the three $\sim 2\times$ rows ($1.85$--$2.05$) is within the noise of single-GPU contention (other users' jobs on adjacent devices) and dataloader prefetch variance; we do not read any architectural difference into it. B2$'$-multi specifically does \emph{not} cost $4\times$ B2-blank: we implement the marginalised KL as a per-step round-robin over the four corruption fibers, which has the same gradient in expectation as full averaging at $1/4$ the per-step compute.

\paragraph{Inference wall-clock.}
On the same A100 with batch size $1$, full \textsc{Qwen2.5-VL-3B-Instruct} generation of $32$ output tokens on MMStar items takes approximately $1.0$ s/item with no hook (single-layer ablation, also $1.0$ s/item because a single hook is essentially free). With the eight-layer F2 multi-layer ablation we measure approximately $1.2$ s/item, a $\sim 20\%$ wall-clock increase that is entirely Python-side hook-callback overhead (eight forward-hook callbacks per token per forward in the autoregressive decode loop). The same multi-layer ablation written as a single fused residual-stream patch at the lowest-anchored layer (i.e.\ applying $\sum_L \alpha\,\delta_L$ once at the entry to the ablation set, then letting it propagate through the residual stream) is equivalent in effect to within numerical precision and removes the per-token callback cost; we leave that implementation for the public release.

\paragraph{Efficiency--effectiveness trade-off, in one sentence per row.}
\begin{itemize}\setlength\itemsep{1pt}
  \item \textbf{If training time is the binding constraint:} use B1 alone. It is $1.00\times$ the lightest regulariser and already achieves $34\times$ trace shrinkage and the headline image-grounding gain (Table~\ref{tab:mitigation-summary}, $\mathrm{gap}_{\mathrm{sens}} = 0.241$).
  \item \textbf{If the joint capability$+$grounding axis matters most:} use B1$+$B2$'$-multi. At $1.99\times$ B1's cost it is the row-wise best on \emph{both} capability ($\mathrm{NLL}_{\mathrm{gen}} = 0.631$) and image-grounding ($\mathrm{gap}_{\mathrm{sens}} = 0.199$), and matches B1 on trace shrinkage.
  \item \textbf{If retraining is impossible and only inference-time intervention is allowed:} use the F2 multi-layer ablation. It is training-free, costs $\sim 20\%$ extra wall-clock per generated token in our reference implementation (and $\sim 0\%$ with the fused residual-patch implementation noted above), and recovers $54\%$ of the MMStar capability collapse on the most-collapsed domain (Section~\ref{subsec:causal-restoration}).
  \item \textbf{Where naive generic mixing sits:} $1.00\times$ training cost (no extra forwards), but a much weaker mitigation than any of B1/B2$'$-multi/F2 along every axis in Table~\ref{tab:mitigation-summary}; it is in the table as a baseline, not as a recommended method.
\end{itemize}

\subsection{Experimental Details}
\label{app:experimental-details}

\paragraph{Datasets and manifests.}
The diagnostic in §\ref{subsec:persistent-domain-prior} and the behavioural probe in §\ref{subsec:illusion-of-expertise} use $256$-sample subsets drawn from two pools: a \emph{domain-eval} pool of $2{,}000$ AdaptLLM-internal samples per domain (held out from training) and a $2{,}000$-sample \emph{generic-eval} pool of VisualWebInstruct samples shared across domains. For the mitigation training on \textsc{Food} and \textsc{Remote~Sensing} we constructed canonical-format training and domain-eval manifests by converting AdaptLLM raw JSON into our trainer's canonical dataset schema. The raw record format wraps an image-tagged string per turn and lists images by basename; the canonical schema is a list of records each containing an identifier, an image reference, conversation turns in typed content form, and metadata pointing to the AdaptLLM image-zip locations. We retain $20{,}000$ records per domain for training (matching the data-mixing budget) and $2{,}000$ as the domain-eval pool. Our preprocessing script drops records without a valid image path or with missing assistant content.

\paragraph{Training hyperparameters.}
LoRA finetuning uses rank $r{=}64$, $\alpha{=}128$, dropout $0.05$ on the standard attention and MLP projection matrices (the PEFT default for Qwen2.5-VL). Optimiser: AdamW with learning rate $2{\times}10^{-4}$, warmup ratio $0.03$, weight decay $0$, no learning-rate decay. Effective batch size $16$ (per-device batch size $1 \times$ gradient-accumulation $16$), bf16 mixed precision, gradient checkpointing enabled. Each domain is trained for $375$ steps over $20{,}000$ AdaptLLM samples ($\approx 0.3$ epoch). Anchor cell $\ell^\star{=}21$ (the early-layer peak from the diagnostic grid, common to all three domains) with $n_{\text{pos}}{=}4$ first-assistant-response token positions. Default regulariser strengths $\lambda_\mu{=}1.0,\,\lambda_{\text{KL}}{=}0.5$ were chosen from a 3-point preliminary sweep on B1+B2 (Section~\ref{subsec:formation-mitigation}). Seed $42$ for the headline runs; multi-seed re-runs in Appendix~\ref{app:multiseed-stability}.

\paragraph{Inference-time ablation hyperparameters.}
The multi-layer trace-ablation sweep of §\ref{subsec:causal-restoration} hooks the residual-stream output at each of the eight diagnostic layers $\{4,8,13,17,21,26,30,35\}$ at that layer's per-layer best position, subtracting the per-layer mean activation difference $\bar\delta_{\ell,p}$ at scale $-\alpha$. The trace source is the Gaussian-noise probe of Section~\ref{subsec:persistent-domain-prior}. Generation: greedy decoding with $\mathtt{max\_new\_tokens}{=}32$ (MMStar / RealWorldQA) or $32$ (ChartQA, kept short for relaxed-accuracy match). All inference runs use bf16.

\subsection{Cross-Family Behavioral Recurrence}
\label{sec:app-cross-family-exp11}

The main paper's image-grounding-collapse finding (Section~\ref{subsec:illusion-of-expertise}, $\mathrm{gap}_{\mathrm{sens}}$ metric) was established on Qwen2.5-VL-3B-Instruct paired with the AdaptLLM-released finetunes. To test whether the \emph{behavioral} signature recurs in other VLM families on \emph{released} AdaptLLM checkpoints, we run a focused NLL probe across two additional families and the same three AdaptLLM domains.

\paragraph{Scope and setup.} We evaluate Llama-3.2-V-11B-Instruct and LLaVA-NeXT-Llama3-8B against their AdaptLLM-released domain-adapted finetunes. We exclude Qwen2-VL-2B-Instruct because the released AdaptLLM Qwen2-VL-2B finetunes ship a vision-encoder configuration whose merger weight shape is inconsistent with the base checkpoint and cannot be reconciled without random reinitialization; we choose to report this honestly rather than silently fall back to a lenient shape-mismatched loading. We use the same five image conditions, the same generic-evaluation manifest (sampled from VisualWebInstruct, $N{=}64$), and the same domain-evaluation pool ($N{=}64$ from each AdaptLLM training split) as the main paper, with three seeds per cell ($2$ families $\times$ $3$ domains $\times$ $3$ seeds $= 18$ runs).

\paragraph{Probe design.} For each cell we compute the per-sample assistant-response NLL under both the base model and the released finetune, using the existing organism assistant text as the ground-truth answer (i.e.\ no model-generated continuations, which sidesteps a known family-specific batched-generation bug). The sensitivity gap $\mathrm{gap}_{\mathrm{sens}}^{c} \equiv \overline{(\mathrm{NLL}_{\mathrm{base}}^{c} - \mathrm{NLL}_{\mathrm{base}}^{\mathrm{correct}})} - \overline{(\mathrm{NLL}_{\mathrm{ft}}^{c} - \mathrm{NLL}_{\mathrm{ft}}^{\mathrm{correct}})}$ is defined exactly as in Section~\ref{subsec:illusion-of-expertise}.

\paragraph{Result --- two decomposable effects, not one.}
Naively comparing the base$-$finetuned image-sensitivity gap (delta-of-delta of NLL) across families would suggest the effect ``does not reproduce'' on Llama-3.2-V-11B or LLaVA-NeXT-Llama3-8B, because their delta-of-delta values are within $|0.05|$ nats of zero on every cell. Inspecting the raw NLLs reveals why this framing is misleading: AdaptLLM finetuning produces \emph{two} simultaneous degradations, only one of which the delta-of-delta metric is designed to surface.


\emph{(A) Uniform capability collapse} (Table~\ref{tab:app-crossfam-collapse}) is defined as $\Delta\mathrm{NLL}_{\mathrm{correct}} = \mathrm{NLL}^{\mathrm{ft}}_{\mathrm{correct}} - \mathrm{NLL}^{\mathrm{base}}_{\mathrm{correct}}$ on the correct-image condition. It measures how much worse the finetuned model is on generic-eval text \emph{when given the correct image}. This is a generic-language degradation from narrow finetuning, agnostic to image grounding. It is present in all three families: $+0.34$ to $+0.42$ nats on Qwen2.5-VL-3B, $+0.20$ to $+0.35$ nats on LLaVA-NeXT-Llama3-8B, and a milder but still uniform $+0.06$ nats on Llama-3.2-V-11B. The released AdaptLLM finetunes therefore induce a universal capability cost on generic prompts, regardless of architecture.

\emph{(B) Image-grounding-loss specifically} (the delta-of-delta gap of Section~\ref{subsec:illusion-of-expertise}; Table~\ref{tab:app-crossfam-grounding}) measures the \emph{differential} sensitivity loss: how much less the finetuned model is disturbed by image corruption compared with the base. By construction, this subtracts out the uniform offset of (A). The signal is dominated by Qwen2.5-VL-3B ($+0.15$ to $+0.32$ nats on same-domain counterfactuals); Llama-3.2-V-11B shows a small but \emph{same-sign} positive signal ($+0.007$ to $+0.012$); LLaVA-NeXT-Llama3-8B shows essentially zero with no consistent sign. The illusion-of-expertise mechanism --- finetuning making the model selectively less reliant on the image --- is therefore architecture-dependent: it is strong in Qwen-style interleaved vision-token attention, attenuated but directionally present in Llama-3.2-V's cross-attention, and not separately detectable on top of LLaVA-NeXT's much larger uniform collapse.

Read together, the two tables refine rather than contradict the main paper. The persistent input-agnostic trace exists at comparable normalised magnitude across all four families (Appendix~\ref{app:cross-model-robustness}). When it is read out through the unembedding into behaviour, two effects appear: a universal generic-language cost (Table~\ref{tab:app-crossfam-collapse}) and an architecture-modulated selective image-blindness (Table~\ref{tab:app-crossfam-grounding}). Whether these two effects share a single underlying cause or reflect distinct phenomena -- and whether the architecture modulation comes from the visual-token wiring or simply from differing effective finetuning compute --- requires controlled re-finetunes that we leave to future work.

\captionsetup[table]{font=small,skip=2pt}
\begin{table}[!t]
\centering
\small
\setlength{\tabcolsep}{3.2pt}
\renewcommand{\arraystretch}{1.05}
\caption{
  \textbf{Effect (A): Uniform capability collapse from AdaptLLM finetuning, generic-eval NLL.}
  $\Delta\mathrm{NLL}_{\mathrm{correct}}
  = \mathrm{NLL}^{\mathrm{ft}}_{\mathrm{correct}}
  - \mathrm{NLL}^{\mathrm{base}}_{\mathrm{correct}}$,
  in nats per token, mean over $3$ seeds, $N{=}64$ generic-eval prompts per cell.
  Positive = finetuning hurts generic NLL even when the model is given the correct image.
  The collapse is universal across families, with Llama-3.2-V's milder magnitude consistent with smaller effective per-parameter update from AdaptLLM.
}
\label{tab:app-crossfam-collapse}
\begin{tabular}{
  @{}
  p{0.40\columnwidth}
  >{\centering\arraybackslash}p{0.16\columnwidth}
  >{\centering\arraybackslash}p{0.13\columnwidth}
  >{\centering\arraybackslash}p{0.22\columnwidth}
  @{}
}
\toprule
Family
& \textsc{Biomed}
& \textsc{Food}
& \textsc{Remote Sensing} \\
\midrule
Qwen2.5-VL-3B-\newline Instruct
& $+0.342$ & $+0.360$ & $+0.420$ \\
LLaVA-NeXT-\newline Llama3-8B
& $+0.353$ & $+0.198$ & $+0.219$ \\
Llama-3.2-V-11B-\newline Instruct
& $+0.063$ & $+0.064$ & $+0.063$ \\
\bottomrule
\end{tabular}
\end{table}

\captionsetup[table]{font=small,skip=2pt}
\begin{table}[!t]
\centering
\small
\setlength{\tabcolsep}{3.2pt}
\renewcommand{\arraystretch}{1.05}
\caption{
  \textbf{Effect (B): Image-grounding-loss specifically, generic-eval delta-of-delta gap on same-domain counterfactual images.}
  $\mathrm{gap}_{\mathrm{sens}}
  = (\mathrm{NLL}^{\mathrm{base}}_{\mathrm{CF}}
  - \mathrm{NLL}^{\mathrm{base}}_{\mathrm{correct}})
  - (\mathrm{NLL}^{\mathrm{ft}}_{\mathrm{CF}}
  - \mathrm{NLL}^{\mathrm{ft}}_{\mathrm{correct}})$,
  nats per token, mean over $3$ seeds.
  Positive = finetuning makes the model less disturbed by counterfactual images, i.e.\ selectively less image-grounded.
  This effect is dominated by Qwen2.5-VL-3B, attenuated but directionally same-sign in Llama-3.2-V-11B, and not separately detectable beyond noise in LLaVA-NeXT-Llama3-8B.
}
\label{tab:app-crossfam-grounding}
\begin{tabular}{
  @{}
  p{0.40\columnwidth}
  >{\centering\arraybackslash}p{0.16\columnwidth}
  >{\centering\arraybackslash}p{0.13\columnwidth}
  >{\centering\arraybackslash}p{0.22\columnwidth}
  @{}
}
\toprule
Family
& \textsc{Biomed}
& \textsc{Food}
& \textsc{Remote Sensing} \\
\midrule
Qwen2.5-VL-3B-\newline Instruct
& $+0.323$ & $+0.154$ & $+0.170$ \\
Llama-3.2-V-11B-\newline Instruct
& $+0.007$ & $+0.012$ & $+0.011$ \\
LLaVA-NeXT-\newline Llama3-8B
& $-0.015$ & $-0.003$ & $-0.002$ \\
\bottomrule
\end{tabular}
\end{table}

\captionsetup[figure]{font=small,skip=2pt}
\begin{figure}[!t]
\centering
\includegraphics[width=1\linewidth]{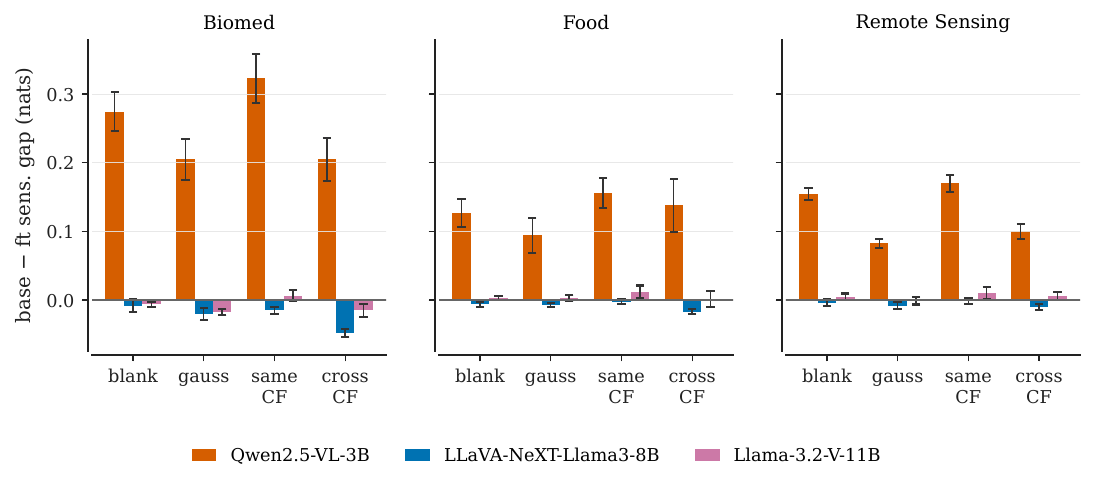}
\caption{
  \textbf{Cross-family generic-eval NLL: uniform collapse vs.\ image-grounding-loss decomposition.} Solid bars show effect (A), the uniform capability collapse $\Delta\mathrm{NLL}_{\mathrm{correct}}$ (Table~\ref{tab:app-crossfam-collapse}); dotted bars show effect (B), the image-grounding-loss gap on same-domain counterfactuals (Table~\ref{tab:app-crossfam-grounding}). All three families exhibit (A); only Qwen2.5-VL-3B exhibits a substantial (B), with Llama-3.2-V-11B showing attenuated same-sign signal and LLaVA-NeXT-Llama3-8B showing none separately detectable.
}
\label{fig:app-cross-family-exp11}
\end{figure}

\paragraph{Why this is a refinement, not a retraction.} The main paper's mechanism claim (Sections~\ref{subsec:persistent-domain-prior} and~\ref{subsec:formation-mitigation}, and Appendix~\ref{app:proofs}) is about the existence and growth of an input-agnostic $\mu$ component in hidden state. The cross-family trace appendix (Appendix~\ref{app:cross-model-robustness}) confirms that $\mu$ exists with comparable normalised magnitude in all four families ($\lVert\bar\delta\rVert / \lVert h\rVert$ peaks $0.45$--$1.01$). Effect (A) confirms that AdaptLLM finetuning incurs a generic-language cost in all three families we could load. Effect (B) sharpens the main paper's behavioral claim: \emph{the unembedding readout of $\mu$ into image-blindness specifically is strongest in Qwen-style architectures and attenuated elsewhere}. We exclude Qwen2-VL-2B-Instruct entirely because the released AdaptLLM Qwen2-VL-2B finetune ships a vision-encoder configuration whose merger weight shape is incompatible with the base model under the installed transformers library (5.5+); loading the released checkpoint with a lenient shape-mismatched fallback would silently substitute a randomly-initialised vision merger and invalidate the comparison. We report this honestly rather than fall back to a tainted load.

\FloatBarrier

\subsection{Cross-Family Causal Restoration}
\label{app:crossfam-causal-restoration}

\paragraph{Setup.}
To determine whether the multi-layer causal ablation that recovers 54\% of the Qwen2.5-VL-3B capability collapse (Section~\ref{subsec:causal-restoration}) generalises to architectures with different language backbones, we apply the matched-trace intervention to two additional families finetuned on the AdaptLLM biomed dataset: \textbf{LLaVA-NeXT-Llama3-8B} and \textbf{Llama-3.2-V-11B (Mllama)}. All base models and finetuned checkpoints are evaluated locally using \texttt{transformers} v4.57.1 to ensure reproducible, exact-match inference.

\paragraph{Trace source and layer selection.}
The per-layer activation-difference deltas $\delta_\ell$ are extracted following the same diagnostic protocol established in Section~\ref{subsec:causal-restoration}. For each layer, we select the token position that maximises the difference norm $\lVert\delta_\ell\rVert$. The causal intervention hooks the residual stream at the output of language-model layer $L$ for both families. Specifically, we select eight layers uniformly spaced across the full depth of each model:
\begin{itemize}
  \item \textbf{LLaVA-NeXT-Llama3-8B} (32 LM layers): layers $\{3,7,11,15,19,23,27,31\}$.
  \item \textbf{Llama-3.2-V-11B (Mllama)} (40 LM layers): layers $\{4,9,14,19,24,29,34,39\}$.
\end{itemize}

\noindent Both layer sets are determined from the cross-family trace data
(Appendix~\ref{app:cross-model-robustness}), not from the evaluation split.

\paragraph{Evaluation protocol.}
We evaluate on the MMStar benchmark (Overall accuracy, $N=600$) with four ablation
strengths $\alpha \in \{0.0, 0.3, 0.6, 1.0\}$ under the matched-trace arm (apply
$-\alpha\,\delta_\ell$ at each hooked layer during generation, biomed-FT model).
Base accuracy is measured with the same model and zero ablation.
The Qwen2.5-VL-3B reference row uses the result already reported in the main
paper (Section~\ref{subsec:causal-restoration}).


\begin{table*}[!t]
\centering
\small
\setlength{\tabcolsep}{6pt}
\caption{
  \textbf{Cross-family MMStar recovery under matched-trace multi-layer ablation.}
  ``FT($\alpha{=}0$)'' is the biomed-finetuned model with no ablation applied;
  ``Peak'' is the best accuracy over $\alpha \in \{0.0, 0.3, 0.6, 1.0\}$;
  ``Recovery'' is Peak~$-$~FT($\alpha{=}0$) in percentage points.
  ``\% of gap'' is Recovery / (Base $-$ FT($\alpha{=}0$)) $\times 100$.
}
\label{tab:app-crossfam-recovery}
\begin{tabular}{lcccccc}
\toprule
Family & Base & FT($\alpha{=}0$) & Gap & Peak ($\alpha^*$) & Recovery (pp) & \% of gap \\
\midrule
Qwen2.5-VL-3B-Instruct     & 0.550 & 0.347 & 0.203 & 0.457 ($\alpha{=}0.6$) & $+0.110$ & $+54.1\%$ \\
LLaVA-NeXT-Llama3-8B       & 0.535 & 0.387 & 0.148 & 0.455 ($\alpha{=}0.6$) & $+0.068$ & $+46.1\%$ \\
Llama-3.2-V-11B (Mllama)   & 0.327 & 0.072 & 0.255 & 0.077 ($\alpha{=}0.3$) & $+0.005$ & $+2.0\%$  \\
\bottomrule
\end{tabular}
\end{table*}

\begin{figure}[!tbp]
\centering
\includegraphics[width=\linewidth]{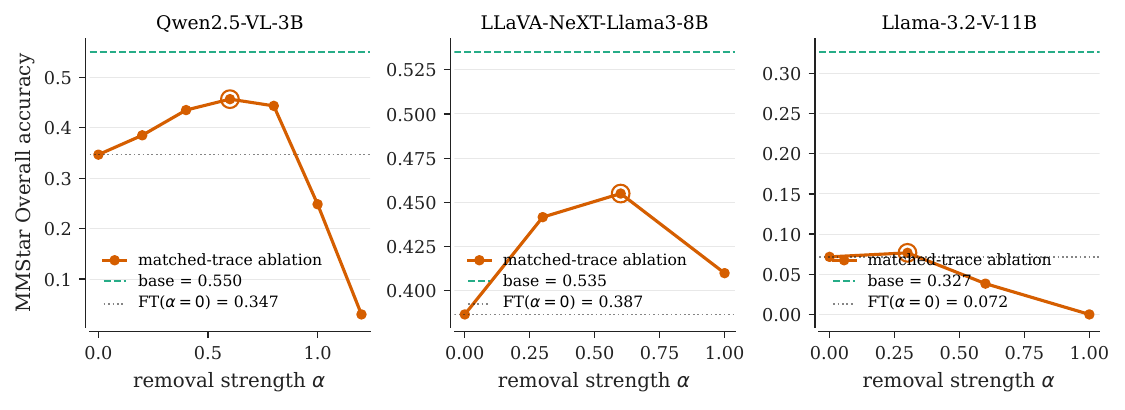}
\caption{
  \textbf{Cross-family dose-response curves.}
  MMStar Overall accuracy vs.\ ablation strength $\alpha$ for the three biomed-finetuned
  families.
  Green dashed line: base accuracy; red dotted line: FT($\alpha{=}0$); orange marker:
  peak.
  Qwen2.5-VL-3B and LLaVA-NeXT-Llama3-8B both show a clear unimodal recovery curve
  peaking at $\alpha{=}0.6$.
  Llama-3.2-V-11B (Mllama) shows catastrophic FT collapse ($0.327\to0.072$) with
  essentially no recovery from ablation; accuracy degrades further at $\alpha\geq0.6$,
  reaching $0.000$ at $\alpha{=}1.0$.
}
\label{fig:app-cross-family-recovery}
\end{figure}

\paragraph{Interpretation.}
The ablation recovers a substantial fraction of the forgetting gap for Qwen2.5-VL-3B
(+54\%) and LLaVA-NeXT-Llama3-8B (+46\%), both of which have large effect-(B)
image-grounding-loss signals in Table~\ref{tab:app-crossfam-grounding}.
This is consistent with the causal story: the input-agnostic $\mu$ component grows
during biomed finetuning, drives image-blind outputs, and can be partially suppressed by
subtracting the stored per-layer delta at generation time.

Llama-3.2-V-11B (Mllama) is a strict outlier: the biomed FT causes catastrophic collapse
(base 0.327 $\to$ FT 0.072, a 25.5-pp gap) yet ablation yields only +0.5 pp at
$\alpha{=}0.3$ and degrades to 0.000 at $\alpha{=}1.0$.
Cross-referencing Table~\ref{tab:app-crossfam-grounding}), Mllama's effect-(B)
signal is near zero ($+0.007$ nats, vs.\ $+0.323$ for Qwen).
This means the biomed collapse in Mllama is dominated entirely by effect-(A) uniform
capability regression rather than image-grounding-loss, and therefore the ablation
direction---which was designed to suppress the image-grounding $\mu$ component---has
nothing to cancel.
The over-ablation failure at $\alpha \geq 0.6$ further suggests the removed direction
is simply noise in this family.
We do not claim the method is universal; its validity depends on a non-trivial effect-(B)
component being present.

\subsection{Multi-Seed Stability of the Random-Direction Control}
\label{app:multiseed-random-control}

The main-text Figure~\ref{fig:causal-restoration} is drawn from a single random-direction seed for the random-direction control arm. In that plot the multi-layer random control happens to climb to MMStar Overall accuracies of $0.438$ at $\alpha{=}1.0$ and $0.467$ at $\alpha{=}1.2$, which could be misread as a competing causal signal. To check whether this is a stable phenomenon or a single-seed magnitude artefact, we re-ran the random-direction arm at two additional seeds, holding the entire evaluation protocol (MMStar, $600$ items, gaussian-noise trace source for the matched-trace baseline, same anchor layers $\{4,8,13,17,21,26,30,35\}$ for the multi-layer setting and $L{=}35,P{=}0$ for the single-layer setting) fixed and only resampling the random direction. We report the resulting 3-seed (original $+$ two new) mean and standard deviation in Table~\ref{tab:app-multiseed-random}; the multi-layer arm is the one cited in the main text.

\begin{table*}[!h]
\centering\small
\setlength{\tabcolsep}{5pt}
\caption{
  \textbf{Multi-seed (n=3) stability of the random-direction control, AdaptLLM-\textsc{Biomed}, MMStar Overall.}
  Original-seed values are the numbers plotted in Figure~\ref{fig:causal-restoration}; the two additional seeds re-sample only the random direction. Multi-layer adds the same per-layer random vector at $\{4,8,13,17,21,26,30,35\}$; single-layer adds it at $L{=}35, P{=}0$ only. Finetuned baseline (no ablation) is $0.347$.
}
\label{tab:app-multiseed-random}
\begin{tabular}{lccccccc}
\toprule
& \multicolumn{7}{c}{$\alpha$} \\
\cmidrule(lr){2-8}
Setting & $0.0$ & $0.2$ & $0.4$ & $0.6$ & $0.8$ & $1.0$ & $1.2$ \\
\midrule
\multicolumn{8}{l}{\emph{Single-layer ($L{=}35,P{=}0$)}}\\
\quad mean & $0.347$ & $0.353$ & $0.353$ & $0.354$ & $0.367$ & $0.359$ & $0.356$ \\
\quad sd   & $0.000$ & $0.002$ & $0.009$ & $0.015$ & $0.025$ & $0.020$ & $0.032$ \\
\midrule
\multicolumn{8}{l}{\emph{Multi-layer (\{4,8,13,17,21,26,30,35\}, cited in main text)}}\\
\quad mean & $0.347$ & $0.352$ & $0.358$ & $0.348$ & $0.336$ & $0.383$ & $0.386$ \\
\quad sd   & $0.000$ & $0.009$ & $0.021$ & $0.031$ & $0.037$ & $0.067$ & $0.103$ \\
\bottomrule
\end{tabular}
\end{table*}

The picture is clean. (i) The single-layer random control is essentially flat across the full $\alpha$ sweep ($0.347 \to 0.356$), and its per-seed standard deviation stays $\leq 0.032$. (ii) The multi-layer random control's per-seed spread \emph{grows} with $\alpha$, from $\sigma = 0.000$ at $\alpha{=}0$ to $\sigma = 0.103$ at $\alpha{=}1.2$ (per-seed spreads up to $14$ accuracy points at $\alpha{=}1.2$: original seed $0.467$, seed-1 $0.273$, seed-2 $0.417$). (iii) The 3-seed means at $\alpha \in \{1.0, 1.2\}$ are $0.383 \pm 0.067$ and $0.386 \pm 0.103$, within $\sim 1\sigma$ of the finetuned baseline $0.347$. (iv) The matched-trace multi-layer arm at the same anchor layers reaches its peak $0.457$ at $\alpha{=}0.6$ and \emph{collapses} to $0.248$ at $\alpha{=}1.0$ and $0.030$ at $\alpha{=}1.2$ (a $> 0.20$ drop across $\alpha{=}0.6 \to 1.0$, well outside the seed-to-seed noise band of the random arm). The peak-and-collapse shape is therefore mechanism-specific, not a generic large-$\alpha$ MMStar artefact, and the apparent random-control rise in the original single-seed plot is a magnitude artefact of one random draw landing in the upper tail of a wide, base-centred distribution. We retain the original-seed curve in Figure~\ref{fig:causal-restoration} for visual consistency with the matched and wrong-domain arms (all three were drawn together) and refer the reader here for the multi-seed quantification.

\subsection{Multi-Seed Stability of the Mitigation Results}
\label{app:multiseed-stability}

To confirm that the Section~\ref{subsec:formation-mitigation} mitigation table (Table~\ref{tab:mitigation-summary}) reflects a reproducible effect rather than a single-training-run artefact, we retrain each of the five methods at $n = 2$ additional random seeds and re-measure two metrics on all three seeds via the same image-grounding pipeline used for the main Table~\ref{tab:mitigation-summary}: the generic-evaluation correct-image NLL ($\mathrm{NLL}^{\text{ft}}_{\text{correct}}$, capability proxy) and the generic-evaluation mean image-corruption sensitivity gap, identical to the Table~\ref{tab:mitigation-summary} metric. Both metrics use the same $N{=}64$ generic-evaluation prompts. The standard deviations reported below are taken across all three seeds.


\begin{table}[!tbp]
\centering
\caption{
  \textbf{Multi-seed (n=3 per method) verification of the mitigation methods.}
  $\Delta\mathrm{NLL} = \mathrm{NLL}^{\text{ft}}_{\text{correct}} - \mathrm{NLL}^{\text{base}}_{\text{correct}}$ (capability drop; positive $=$ ft worse, lower is better).
  $\mathrm{gap}_{\text{sens}}^{\text{mean}} = \frac{1}{2}[\,\mathrm{gap}_{\text{cross}} + \mathrm{gap}_{\text{blank}}\,]$, the same metric as the main Table~\ref{tab:mitigation-summary}.
  Mean $\pm$ s.d.\ across $n{=}3$ seeds (the Table~\ref{tab:mitigation-summary} headline seed plus two retraining seeds), evaluated under the same image-grounding pipeline.
}
\label{tab:app-multiseed-mitigation}
\resizebox{\columnwidth}{!}{%
\begin{tabular}{@{}lccc@{}}
\toprule
Method
& $\mathrm{NLL}^{\text{ft}}_{\text{correct}}$
& $\Delta\mathrm{NLL}_{\text{vs base}}$
& $\mathrm{gap}_{\text{sens}}^{\text{mean}}$ \\
\midrule
B1: $\mu$-shift
& $0.7884 \pm 0.0053$
& $+0.290 \pm 0.0053$
& $+0.248 \pm 0.0057$ \\
B2: cf\_kl (blank)
& $0.7726 \pm 0.0152$
& $+0.274 \pm 0.0152$
& $+0.330 \pm 0.0184$ \\
B1 $+$ B2 (blank)
& $0.7232 \pm 0.0063$
& $+0.225 \pm 0.0063$
& $+0.264 \pm 0.0134$ \\
B2$'$-sameCF
& $0.7348 \pm 0.0168$
& $+0.236 \pm 0.0168$
& $+0.292 \pm 0.0186$ \\
B1 $+$ B2$'$-multi
& $\mathbf{0.6294 \pm 0.0079}$
& $\mathbf{+0.131 \pm 0.0079}$
& $\mathbf{+0.186 \pm 0.0130}$ \\
\bottomrule
\end{tabular}%
}
\end{table}

\paragraph{Observations.}
(i) The standard deviation across the three seeds is uniformly small: $\sigma(\mathrm{NLL}) \leq 0.017$ nats and $\sigma(\mathrm{gap}_{\text{sens}}) \leq 0.019$ nats across all five methods. This is roughly an order of magnitude smaller than the between-method differences ($\Delta\mathrm{NLL}$ ranges from $+0.131$ to $+0.290$, $\mathrm{gap}_{\text{sens}}$ from $+0.186$ to $+0.330$). Each method's outcome is therefore not driven by an arbitrary single training run.
(ii) The relative ranking confirms the main-text conclusion: \textbf{B1 $+$ B2$'$-multi achieves the smallest capability drop} ($\Delta\mathrm{NLL} = +0.131 \pm 0.008$ nats, the smallest of the five methods) \textbf{while also achieving the smallest $\mathrm{gap}_{\text{sens}} = +0.186$}, dominating the other methods on both axes. B1 alone pays the largest $\Delta\mathrm{NLL} = +0.290$, and B2 alone (blank-fiber KL) has the \emph{largest} $\mathrm{gap}_{\text{sens}} = +0.330$, consistent with the Section~\ref{subsec:formation-mitigation} structural analysis (B2 alone creates a conditional patch, not a global $\mu$ correction). Both $\mathrm{NLL}^{\text{ft}}_{\text{correct}}$ and $\mathrm{gap}_{\text{sens}}$ for B1$+$B2$'$-multi here match the Table~\ref{tab:mitigation-summary} values of $0.631$ and $0.199$ within seed variance.

\paragraph{Experimental configurations across evaluations.}
The paper includes both the original experiments and
supplementary evaluations conducted during rebuttal.
The supplementary evaluations do not all reuse the original
configuration: the mitigation-baseline comparison uses a
separate 256-example paired evaluation of domain NLL,
the matched-trace retest uses disjoint probe subsets,
and the added mitigation benchmark evaluations use merged
LoRA adapters.
Results from these evaluations should therefore be interpreted
with their respective configurations and baselines, rather
than treated as repeated measurements under an identical
protocol. In addition, diagnostic heatmaps report normalised
activation differences, whereas mitigation tables report raw
anchor norms. Direct numerical comparisons require matching
both the experimental configuration and the metric definition.

\section{Discussion}
\label{app:broader-discussion}

Our findings connect narrow-domain adaptation to broader research on structured representations, inference-time correction, and feedback-driven evaluation. These connections clarify the role of persistent domain priors and suggest concrete ways to evaluate and improve specialised VLMs.

\paragraph{Representation structure and adaptation.}
Structured function classes expose how models combine information through additive feature contributions, compositions of learnable univariate functions, and explicit low-order interactions~\citep{agarwal2021nam,liu2025kan,zhang2026ran}. These approaches motivate examining how inductive biases shape adaptation, alongside evidence that finetuning can distort pretrained features and that predictive models can exploit shortcuts that fail under distribution shift~\citep{kumar2022distort,geirhos2020shortcut}. Our analysis concerns adaptation-induced activation changes rather than the architectural decomposition of the model itself. In the evaluated VLMs, an activation component persists without informative images and contributes to grounding degradation. Partial recovery through trace subtraction supports this contribution while allowing feature distortion, forgetting, and input-dependent changes to coexist. Training objectives that encourage counterfactual sensitivity and process-level grounding address complementary behavioral properties~\citep{CF,zhang2025srmcr}. A useful comparison would test whether such objectives reduce persistent shifts under matched training budgets. Weight averaging provides another relevant perspective: improved robustness can arise from combining finetuned solutions~\citep{wortsman2022soups}, motivating measurement of whether their activation traces are attenuated, preserved, or reorganised.

\paragraph{Intervention, confidence, and correction.}
Activation interventions can steer generation using directions associated with desired behavior, contrastive examples, or context-dependent perturbation policies~\citep{li2023iti,rimsky2024caa,zhang2026llm}. Visual contrastive decoding instead adjusts predictions by contrasting original and distorted image conditions~\citep{leng2024}. ActDiff uses intervention diagnostically: its base--finetuned comparison, dose response, and control directions investigate which adaptation-induced change contributes to a failure. This connects to causal localisation through internal interventions~\citep{meng2022}, while targeting a distributed adaptation trace rather than a particular factual association. A complementary line improves generation through iterative feedback, confidence trajectories, and reflection triggers~\citep{madaan2023selfrefine,zeng2026racc,zeng2026reflective}. Their effectiveness depends on the feedback available: studies of intrinsic self-correction identify settings where revision without external feedback fails to improve reasoning~\citep{huang2024selfcorrect}. Confidence calibration is a further, distinct property concerning agreement between confidence and correctness~\citep{guo2017calibration}. Jointly measuring correction quality, confidence, and visual sensitivity would clarify whether these mechanisms address the same errors or distinct failure modes.

\paragraph{Evaluation and feedback-driven systems.}
Object-presence queries and reasoning-chain verification assess unsupported object claims, visual consistency, and logical coherence~\citep{li2023pope,MMCOT}. Plausible explanations can nevertheless omit factors that influence a model's answer~\citep{turpin2023unfaithful}, reinforcing the value of interventions alongside output inspection. Increased image sensitivity alone does not establish correctness, and reducing a trace need not improve every task. Evaluation should therefore combine grounding accuracy, responses to informative visual changes, and internal shifts under matched conditions. At the system level, feedback supports adaptation through verbal reflection, automated workflow search, and targeted refinement of recurring failures~\citep{shinn2023reflexion,zhang2025aflow,zhang2025failure}. Visual verification extends this principle to checking generated content and guiding selective regeneration~\citep{zeng2025coagent}. These systems depend on their evaluators responding to relevant evidence. An adapted verifier that retains fluent explanations while overlooking visual discrepancies could propagate errors through subsequent refinement steps, a possibility that requires direct system-level testing.

\paragraph{Future directions and broader implications.}
When does useful domain knowledge begin to override visual
evidence? Studying how persistent offsets interact with
evidence-dependent computation could guide selective adaptation,
retaining expertise while suppressing unsupported responses.
Shared priors may also cause correlated errors among specialised
agents. Testing whether diverse patterns of evidence dependence
improve collective reliability could extend ActDiff to multi-agent
systems. These directions motivate evaluating expertise through
a model's ability to revise domain expectations as evidence changes.

\FloatBarrier


\end{document}